\documentclass[10pt,twocolumn,letterpaper]{article}

\usepackage{iccv}              

\usepackage[accsupp]{axessibility} 
\usepackage{xcolor}
\usepackage{bibentry}

\usepackage{amsmath,amssymb,amsfonts}
\usepackage{adjustbox}
\usepackage{dsfont}
\usepackage{diagbox}
\usepackage{multirow}
\usepackage{mathdots}
\usepackage{mathrsfs}
\usepackage{pifont}

\definecolor{iccvblue}{rgb}{0.21,0.49,0.74}
\usepackage[pagebackref,breaklinks,colorlinks,allcolors=iccvblue]{hyperref}

\def\paperID{10194} 
\def\confName{ICCV}
\def\confYear{2025}

\title{Background Invariance Testing According to Semantic Proximity}

\author{Zukang Liao\\
University of Oxford\\
{\tt\small zukang.liao@eng.ox.ac.uk}
\and
Min Chen\\
University of Oxford\\
{\tt\small min.chen@oerc.ox.ac.uk}
}

\begin{document}
\maketitle
\begin{abstract}
In many applications, machine-learned (ML) models are required to hold some invariance qualities, such as rotation, size, and intensity invariance. Among these, testing for background invariance presents a significant challenge due to the vast and complex data space it encompasses. To evaluate invariance qualities, we first use a visualization-based testing framework which allows human analysts to assess and make informed decisions about the invariance properties of ML models. We show that such informative testing framework is preferred as ML models with the same global statistics (e.g., accuracy scores) can behave differently and have different visualized testing patterns. However, such human analysts might not lead to consistent decisions without a systematic sampling approach to select representative testing suites. In this work, we present a technical solution for selecting background scenes according to their semantic proximity to a target image that contains a foreground object being tested. We construct an ontology for storing knowledge about relationships among different objects using association analysis. This ontology enables an efficient and meaningful search for background scenes of different semantic distances to a target image, enabling the selection of a test suite that is both diverse and reasonable. Compared with other testing techniques, e.g., random sampling, nearest neighbors, or other sampled test suites by visual-language models (VLMs), our method achieved a superior balance between diversity and consistency of human annotations, thereby enhancing the reliability and comprehensiveness of background invariance testing.
\end{abstract}    
\section{Introduction}
\label{sec:intro}There are a variety of invariance qualities associated with machine-learned models.
Testing these invariance qualities enables us to evaluate the robustness of a model in its real-world application, where the model may encounter variations that do not feature sufficiently in the training and testing data. Testing also allows us to observe possible biases or spurious correlations that may have been learned by a model \cite{spurious_aaai} and to anticipate whether the model can be deployed in other application domains \cite{domain_generalisation}.
This work is concerned with background invariance testing -- a relatively challenging type of testing. 

\begin{figure}[t]
\centering
\includegraphics[height=13.5mm]{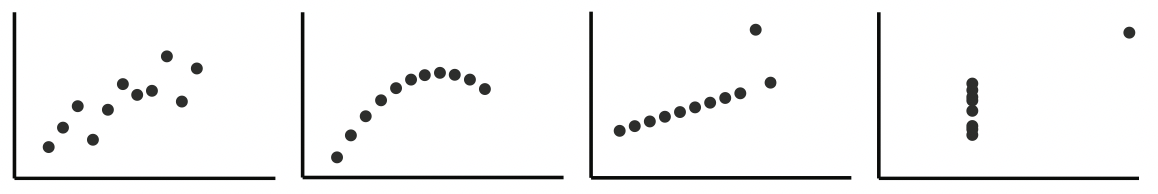}
\caption{The four sets of points, i.e., the Anscombe's quartet \cite{Anscombe}, have exactly the same statistical measures, e.g., mean, standard deviation, correlation, etc. However, they are differently distributed. 
Visualization-based approaches are often more informative than statistical scores for illustrating data distributions.
}
\label{fig:anscombe}
\end{figure}

\begin{figure}[t]
    \centering
    \includegraphics[height=27mm]{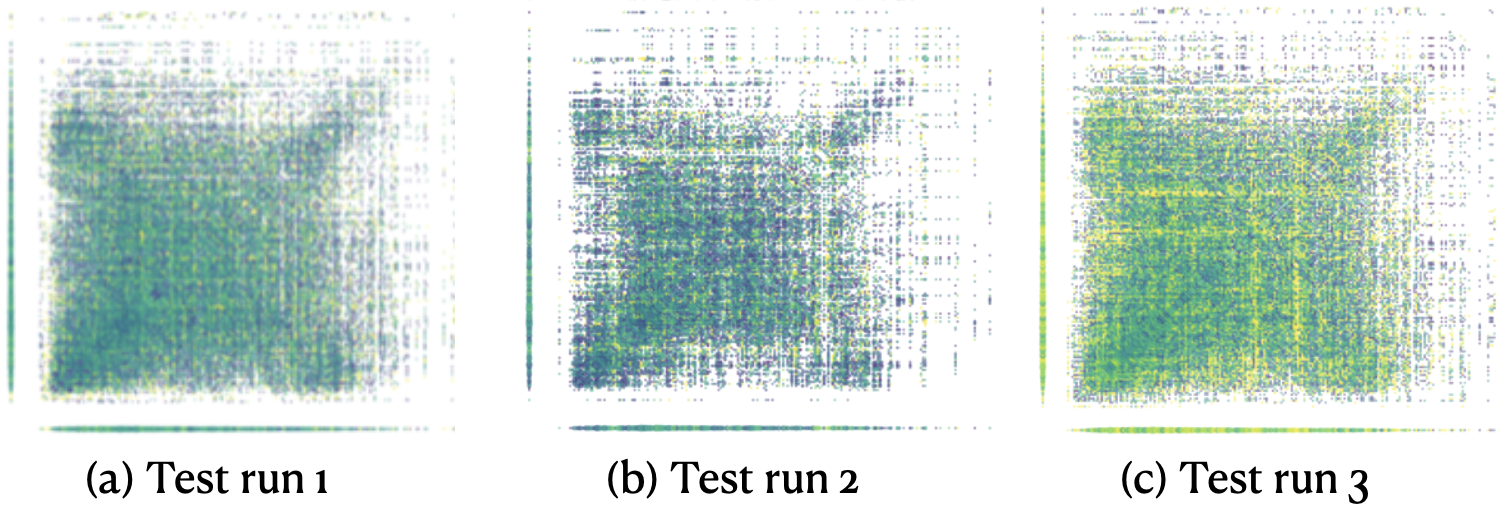}
    \caption{Random sampling leads to inconsistent visual representations across different testing runs, making visualization-based testing frameworks or human-centric testing methods inconsistent. We show the proposed testing framework leads to more consistent testing patterns in Appendix B4.}
    \label{fig:VarianceMatrix}
\end{figure}

\begin{figure*}
    \centering
    \includegraphics[width=0.85\linewidth]
    {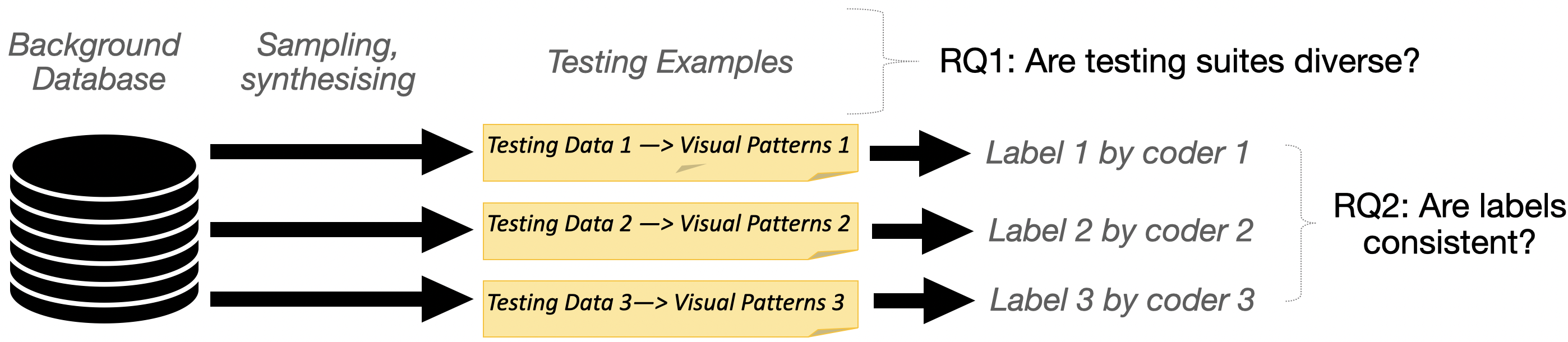}
    \caption{Research questions: to utilize informative visualization-based techniques for background invariance testing, this paper focused on: 1) are the selected testing examples diverse, and 2) are the resultant human decisions (based on the visual patterns) consistent.}
    \label{fig:rq}
\end{figure*}

When considering invariance qualities of an ML model, most existing works reported a single averaged worst-case accuracy, e.g., \cite{bgchallenge, fuzzaugmentation}. However, as shown in Figure \ref{fig:anscombe}, typical statistical measures, e.g., mean, cannot informatively characterize how the data are distributed. In machine learning (ML), while averaged accuracy scores can provide an overall statistical indication of the invariance quality, they do not support more detailed analysis such as whether the level of robustness or biases is acceptable in an application by taking into account the probabilities of the variants, for example different background scenes. To transform the problem of invariance testing from simply reporting an accuracy score into a more informative multi-factor decision-making process, recent work \cite{ml4ml_invariance, ExploringLandscape} conducted testing for basic invariance qualities, e.g., rotation etc, based on visual matrices (formed using all testing results). In this work, we show that ML models with the same accuracy and worst-case accuracy score can exhibit different visualized testing patterns. Therefore, we believe a visualization-based testing framework is preferred to conduct more informative tests.

However, visualization-based testing frameworks often rely on human judgment of visualized patterns. 
When working with a vast data space, selecting only a small subset for evaluation without a systematic sampling strategy can result in inconsistent test-example selections across different runs. This inconsistency leads to shifting visual patterns and unstable human judgments, as illustrated in Figure \ref{fig:VarianceMatrix}. 
Conversely, a sampling strategy that favors a specific type of test case may yield more consistent human judgments but at the cost of reduced diversity in the test suite. To effectively leverage visualization-based testing (more informative) for background invariance, this paper aims to balance the trade-off between test data diversity and judgment consistency (as shown in \ref{fig:rq}). 
Our contributions are:

\begin{itemize}
    \item [a.] We qualitatively confirm that the visualization-based testing method is more informative by showing that ML models with the same averaged worst-case accuracy score can behave differently (i.e., different visual patterns).
    \item [b.] We qualitatively confirm that the visualization-based testing method suffers from trade-offs between diversity of testing suites and consistency between human decisions.
    \item [c.] To overcome the trade-off, we introduce an algorithm to search for $n$ desired background scenes based on the semantics encoded in each original image using association ontology.
    \item [d.] We quantitatively prove that our testing approach is the most balanced between diversity (\emph{recall}) and consistency (\emph{precision}) with the highest f1 score, compared with other testing methods.
    \item [e.] We show that the proposed background invariance testing based on visual representations can be fully automated.
\end{itemize}

\section{Related Works}
\label{sec:related_work}
Invariance qualities of ML models have been studied for a few decades. In recent years, invariance testing has become a common procedure in invariant learning \cite{invariant_risk,water_bird,invariant_learning}. Among different invariance qualities, background invariance is attracting more attention. In the literature, several types of variations were introduced in background invariance testing, e.g., by replacing the original background with random noise, color patterns, and randomly selected background images. In this section, firstly, we introduce existing attempts, followed by an introduction to other techniques that are used in this work to help conduct background invariance testing.

Rosenfeld, et al. \cite{noise_bgtest} tested object detection models by transforming the original background to random noise or black pixels. They reported that all tested models failed to perform correctly at least in one of their testing cases. Similarly,
others, e.g., \cite{random_erase,aaai_occlusion1,aaai_occlusion2}, replaced parts of the images with black or gray pixels for foreground invariance testing. \cite{2004scenetest} noticed that the association between a foreground object and its background scene affected object recognition and described such association as ``consistency''.
Lauer and Cornelissen
\cite{color_bgtest} tested different models with consistent and inconsistent backgrounds, while using the term ``semantically-related'' to describe consistent association.
In particular, they used color texture to replace the original background of the target image and controlled the inconsistency using a parameterized texture model \cite{color_texture}.
Several researchers experimented with swapping background scenes in studying background invariance, e.g., \cite{2004scenetest}.
Xiao, et al. \cite{bgchallenge} provided the Background Challenge database by overlaying a foreground object to all extracted backgrounds from other images. To prepare models (to be tested), they also provided a smaller version of ImageNet with nine classes (IN9). In this work, we train a small repository of models on IN9, i.e., the models being tested in this work were trained for image classification.

\begin{figure*}[t]
    \centering
    \includegraphics[width=\linewidth]{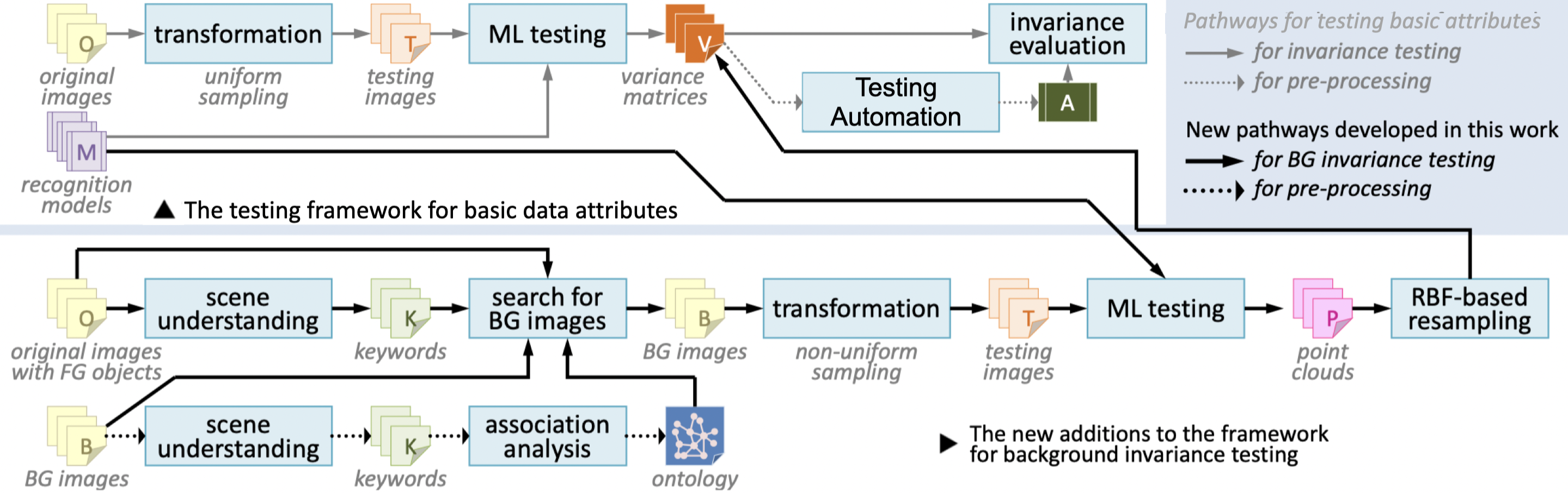}
    \caption{The upper part of the figure shows the invariance testing framework for simple data attributes, e.g., rotation, where the transformations for invariance testing are uniformly sampled. As the transformations for background invariance testing cannot be easily sampled in a consistent way, we introduce a new sub-workflow (lower part) with an additional set of technical components to enable non-uniform sampling of such transformations. This sub-workflow is detailed in Section \ref{sec:mehod} Methodology. All the trained models and datasets are available at \url{https://github.com/Zukang-Liao/background_invariance_testing}.}
    \label{fig:Framework}
\end{figure*}

To test deep models, an ideal testing suite should trigger as many neurons as possible. Pei et al. \cite{deepxplore} defined the percentage of neurons triggered by a testing suite as the \emph{neuron coverage rate} of the testing suite. Recent studies utilize coverage-based fuzzing techniques to find a testing suite that triggers more potential ``bugs" for an ML model, e.g., TensorFuzz \cite{tensorfuzz} and DeepHunter \cite{deephunter}. In this work, we adopt the most commonly used neuron coverage rate \cite{deepxplore} to evaluate the synthesized testing images; Higher neuron coverage rates indicate more diverse testing images \cite{surprise}. For these reasons, in this work, we use neuron coverage rate to indicate how diverse and comprehensive a testing suite is.

More recent works on invariance testing found that averaged accuracy scores are not informative enough to judge the performance of ML models \cite{xai_oppor_challenge_survey, xai_survey}. Instead, visualized representations are becoming more popular for conducting more informative tests and analyses. For example,
Engstrom et al. \cite{ExploringLandscape} used 3D heatmaps to analyze the translation/rotation invariance qualities.
Liao et al. \cite{ml4ml_invariance} used visualization to depict different model behaviors for which statistics cannot.
Liao and Cheung \cite{sparse_cikm} showed that analyzing the invariance qualities based on visualized patterns can achieve a higher inter-rater reliability (IRR) score than many NLP tasks, confirming the plausibility of conducting invariance testing based on visual patterns. In this work, we show that our visualization-based background invariance testing framework can lead to a satisfactory IRR score with the assistance of our novel technical components, e.g., ontology built on association analysis.

\section{Definition, Overview, and Motivation}

Let $\mathbf{x}_i$ be the $i^{th}$ image in a dataset $D$, $o_i$ be the foreground object in $\mathbf{x}_i$, and $\mathbf{M}$ be an ML model trained to recognize or classify $o_i$ from $\mathbf{x}_i$. In general, the invariance quality of $\mathbf{M}$ characterizes the ability of $\mathbf{M}$ to perform consistently when a type of transformation is applied to $\mathbf{x}_i$. For example, one may apply a sequence of rotation transformations $\mathbf{y}_{i,j} = R(\mathbf{x}, j^{\circ}), \; j=0, 1, \ldots$, and test $\mathbf{M}$  with the newly transformed/generated testing images of $\mathbf{y}_{i,j}$. When the testing results can easily be sampled and organized, the visual patterns can facilitate detailed human analysis, e.g., whether the level of robustness or biases is acceptable when taking into account the probabilities of the variants.

The background invariance quality characterizes the ability of $\mathbf{M}$ in recognizing $o_i$ when it is with different backgrounds. Hence the transformations of $\mathbf{x}_i$ involve the replacement of the original background in $\mathbf{x}_i$ with different background scenes $\mathbf{b}_1, \mathbf{b}_2, \ldots, \mathbf{b}_n$. The transformations:
\begin{equation}\label{eq:BG-Transform}
    \mathbf{y}_{i,j} = \textit{Mask}_i \circledast x_i + (1-\textit{Mask}_i) \circledast \mathbf{b}_j, \; j=1,2,\ldots,n
\end{equation}
where $\textit{Mask}_i \circledast x_i=o_i$. However, selecting a meaningful and suitable testing suite from an enormous data space is difficult. Therefore, background invariance testing is often carried out based on random selection, resulting in meaningless visual representations and unreliable human judgments (Figure \ref{fig:VarianceMatrix}). If we can find a way to consistently produce visual representations for the testing results from background transformations, we should be able to conduct visual analysis and the judgment on the background invariance qualities should be consistent and reliable. This motivates us to address the following challenges:

\begin{enumerate}
    \item to have an effective way to search for background scenes that will be distributed appropriately to form meaningful visualized testing results.
    \item to show that the selected background scenes are diverse, broad, and representative.
    \item to show that the visual representations are meaningful and the judgments based on them are reliable.
    \item to show that such testing procedure can also be automated, therefore it becomes less labor-intensive and more time-efficient.
\end{enumerate}

\begin{figure*}[t]
    \centering
    \includegraphics[width=\textwidth]{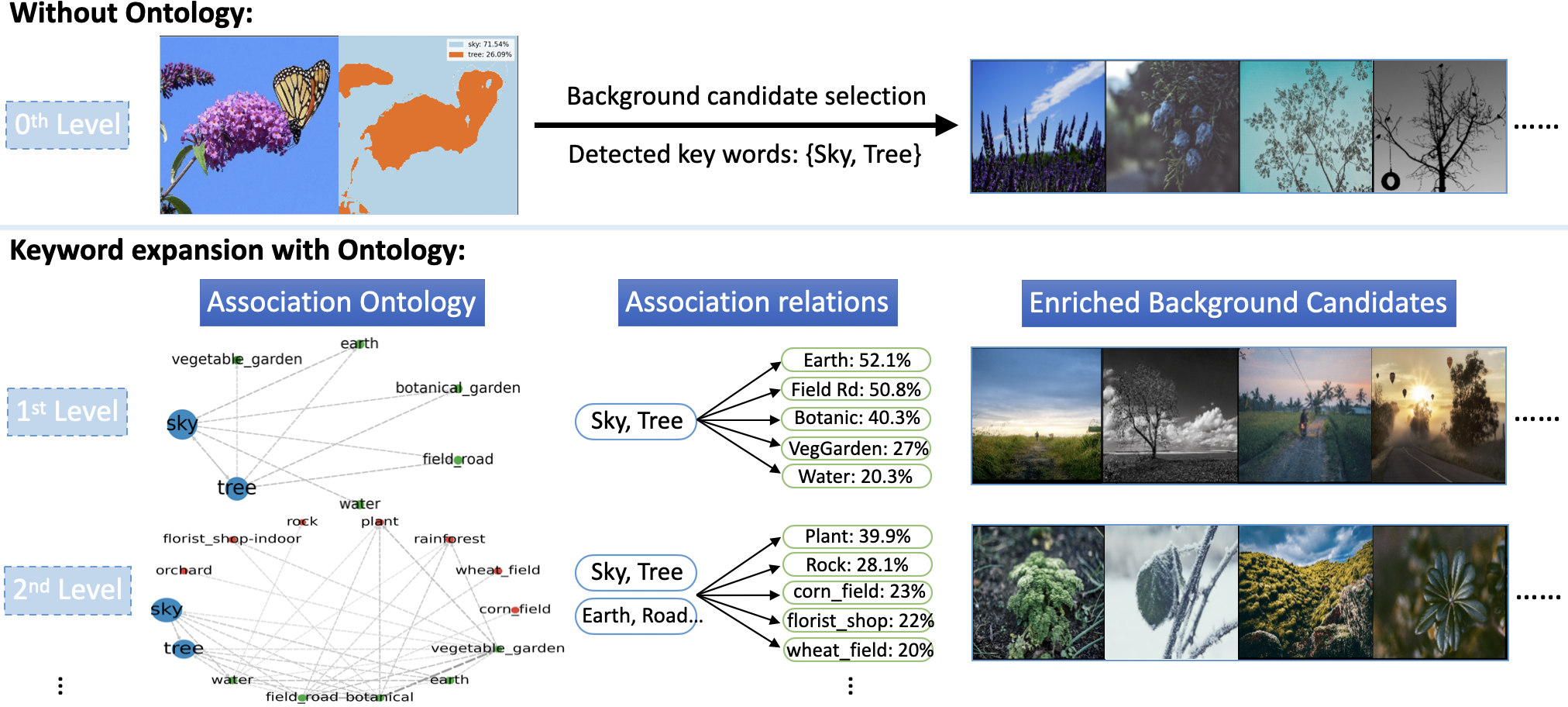}
    \caption{An example image has only two keywords detected by a pre-trained scene understanding model, namely $\{sky, tree\}$. Using an ontology, more keywords can be discovered iteratively, increasing the number and diversity of background scenes.}
    \label{fig:Ontology}
\end{figure*}

In this work, we introduce a number of technical components (lower part of Figure \ref{fig:Framework}) to address the aforementioned challenges.
Once these challenges are addressed, background invariance testing can be improved from reporting a single average accuracy score to a consistent and reliable judgment based on tailored visual representations.

\section{Methodology}
\label{sec:mehod}
In this section, we follow the pathways in the lower part of Figure \ref{fig:Framework} to describe a series of technical solutions for enabling background invariance testing with non-uniform sampling of the transformations of the original images.

\subsection{To Obtain Detected and Expanded Keywords} 
In prior works, background candidates were sampled randomly, leading to inconsistent visual representations. In this work, we introduce a systematic sampling approach to obtain background testing candidates. Firstly, we use a scene understanding model and association analysis to build an ontology. We then use the ontology to retrieve indirectly relevant keywords to the original images. Finally, with a set of detected and expanded keywords, keyword-based sampling is used to obtain background testing candidates.
%
\subsubsection{Detected Keywords (Scene Understanding)}
From each testing image, a scene understanding model identifies a set of objects that are recorded as a set $K_a$ of keywords. Using the keywords extracted from each image, multiple keywords can be identified for most images, but in many cases, fewer than 3 keywords were detected. We list the statistics in Figure \ref{fig:nbitems} and show some examples in Figure \ref{fig:scene_item} in the Appendix A. To select a suitable (non-random) testing suite that contains rich semantics, it is desirable to consider not only the original keywords, but also other keywords that are related to the detected keywords.

\subsubsection{Expanded Keywords: Association Ontology}
\label{sec:ontology}
%
To consider indirectly relevant keywords to the original testing image, we built an ontology using association analysis, i.e., Apriori algorithm \cite{Apriori} and Frequent Pattern Growth algorithm \cite{fptree}.
Given a set of all possible keywords $K_{\text{all}}$ that can be extracted from all images in a dataset $\mathbb{B}$,
the level of association between two keywords $k_a$ and $k_b$ can be described by \emph{support} and \emph{confidence}. For three itemsets: $s_a = \{k_a\}$, $s_b = \{k_b\}$, and $s_{ab} = \{k_a, k_b\}$, 
the \emph{support} for the itemset $s_{ab}$ is defined as:
\[
\emph{support}(s_{ab}) = \frac{\text{number of images where } s_{ab} \text{ is present}}{\text{total number of images}}
\]
\noindent which indicates the co-occurrence rate of $k_a$ and $k_b$.

\begin{figure*}[t]
    \centering
    \includegraphics[width=\textwidth]{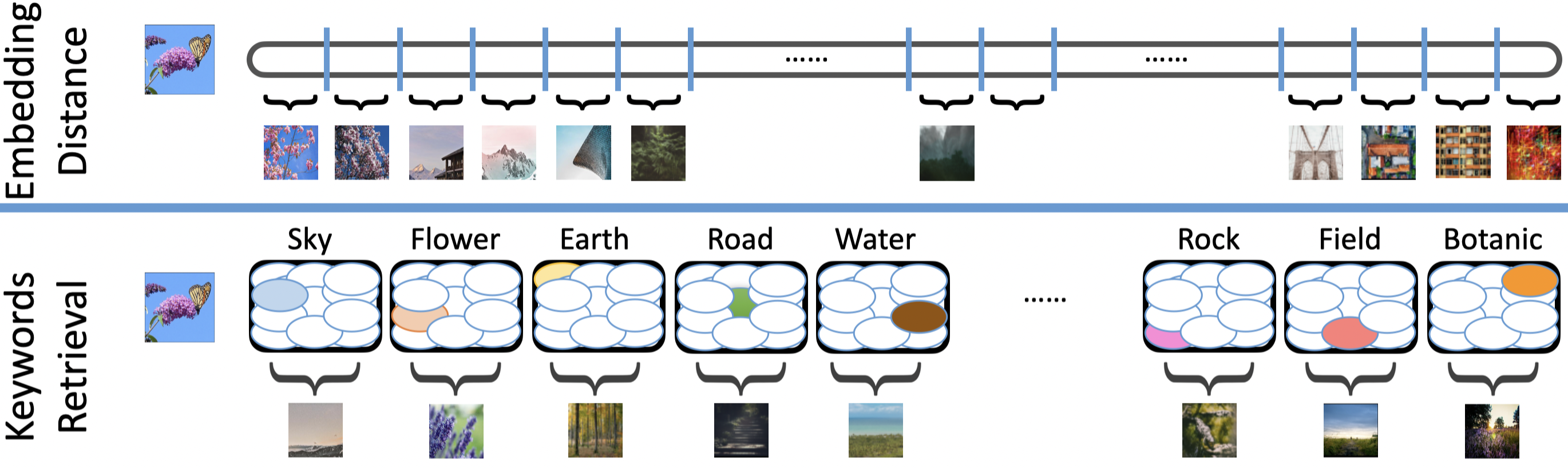}
    \caption{We can sample background scenes based on: 1) similarity/distance metrics, or 2) keywords. When sampling one instance from each subspace, we choose the first one leads to a higher realness score defined by \cite{dreamsim} than a threshold.}
    \label{fig:bg_sample}
\end{figure*}

An association rule from one itemset to another, denoted as $\exists s_a \rightarrow \exists s_b$, is defined as \emph{confidence}:
\begin{equation}\label{eq:Confidence}
    \emph{confidence}(\exists s_a \rightarrow \exists s_b) = \frac{\emph{support}(s_a\bigcup s_b)}{\emph{support}(s_a)}
\end{equation}
\noindent which indicates the confidence level about the inference that if the object of keyword $k_a$ appears in a scene, the object of keyword $k_b$ could also appear in such a scene.
Similarly, we can compute \emph{confidence}$(\exists s_b \rightarrow \exists s_a)$. For some non-hierarchical specific keyword, the value of \emph{support}$(s_{1,2})$ is usually tiny, and is more easily changed by the increase of images in the repository, the introduction of more keywords, and the improvement of scene understanding techniques. We therefore use \emph{confidence} as the weights (directed edges) in our ontology.

In the ontology, the shortest path between two keywords indicates the level of association between them, typically facilitating two measures, i) the number of edges along the path (i.e., hops) and ii) an aggregated weight, e.g., $\prod_i^h w_{i=1}$ or $\min(0, w_1-\sum_{i=2}^h (1-w_i)^{\alpha_i} (\alpha_i \geq 1))$. As illustrated in Figure \ref{fig:Ontology}, nodes represent keywords, and an edge between two nodes indicates that two keywords have been detected from the same image at least once. The weight on the edge indicates how strong is the association between the two keywords. The ontology is typically constructed in a pre-processing step by training association rules using the extracted keywords for all images in a dataset.

A set of keywords $K_x$ extracted by a scene understanding model can be used to search for background scenes with at least one of the matching keywords $k \in K_x$. When there are many keywords in $K_x$, search based on original keywords can work very well. However, as exemplified in Figure \ref{fig:Ontology}(top), when an image has only two keywords, the search will likely yield a small number of background scenes, undermining the statistical significance of the test.

To address this issue, we expand the keyword set $K_x$ by using the ontology that has acquired knowledge about keyword relationships in the preprocessing stage. As illustrated in Figure \ref{fig:Ontology}, the initial set $K_x$ has keywords [sky, tree].
The ontology shows that $\{$Sky, Tree$\}$ are connected to $\{$Earth, Field Road, Botanic Garden, Vegetable Garden, Water$\}$, which form the level 1 expansion set $E_{1,x}$. Similarly, from $E_1$, the ontology helps us to find the level 2 expansion set $E_{2,x}$, and so on.
The set of all keywords after $i$-th expansion is:
\begin{equation}\label{eq:OntologyLevel}
    \text{OL}_{x}[i] = K_x \cup \biggl( \bigcup_{j=1}^{i} E_{i,x} \biggr)
\end{equation}

\subsection{To Synthesize/Generate Testing Images} 
For an original image, with a set of keywords (detected and expanded), one can synthesize testing images by i) generative blending, or ii) simple background replacement Eq.\,\ref{eq:BG-Transform}. In this work, we use the latter to avoid unwanted foreground objects. We guarantee that no foreground objects will appear in any background scene by carefully selecting the dataset of the background candidates.

\subsubsection{Background Scenes Sampling}
\label{sec:bg_sampling}
Given a target image $\mathbf{x}$, to test if an ML model is background-invariant, we need to sample a set of background scenes that can be used to replace the original background in $\mathbf{x}$ while maintaining the foreground object $o$. 
As shown in Figure \ref{fig:bg_sample}, to replace random sampling, we can sample background scenes based on: a$_1$) cosine/l2 distance between embeddings of the original image and testing images, or a$_2$) keywords. When (randomly) sampling one instance from a subspace defined by b$_1$) a distance interval (bin), or b$_2$) background scenes containing a certain keyword, we run the Dreamsim \cite{dreamsim} model to select the first background scene that leads to a testing image with a higher realness score than a threshold. In Section 5, we quantitatively show that keyword-based sampling is the most balanced between diversity and reliability.

\begin{figure*}[t]
    \centering
    \includegraphics[width=\linewidth]{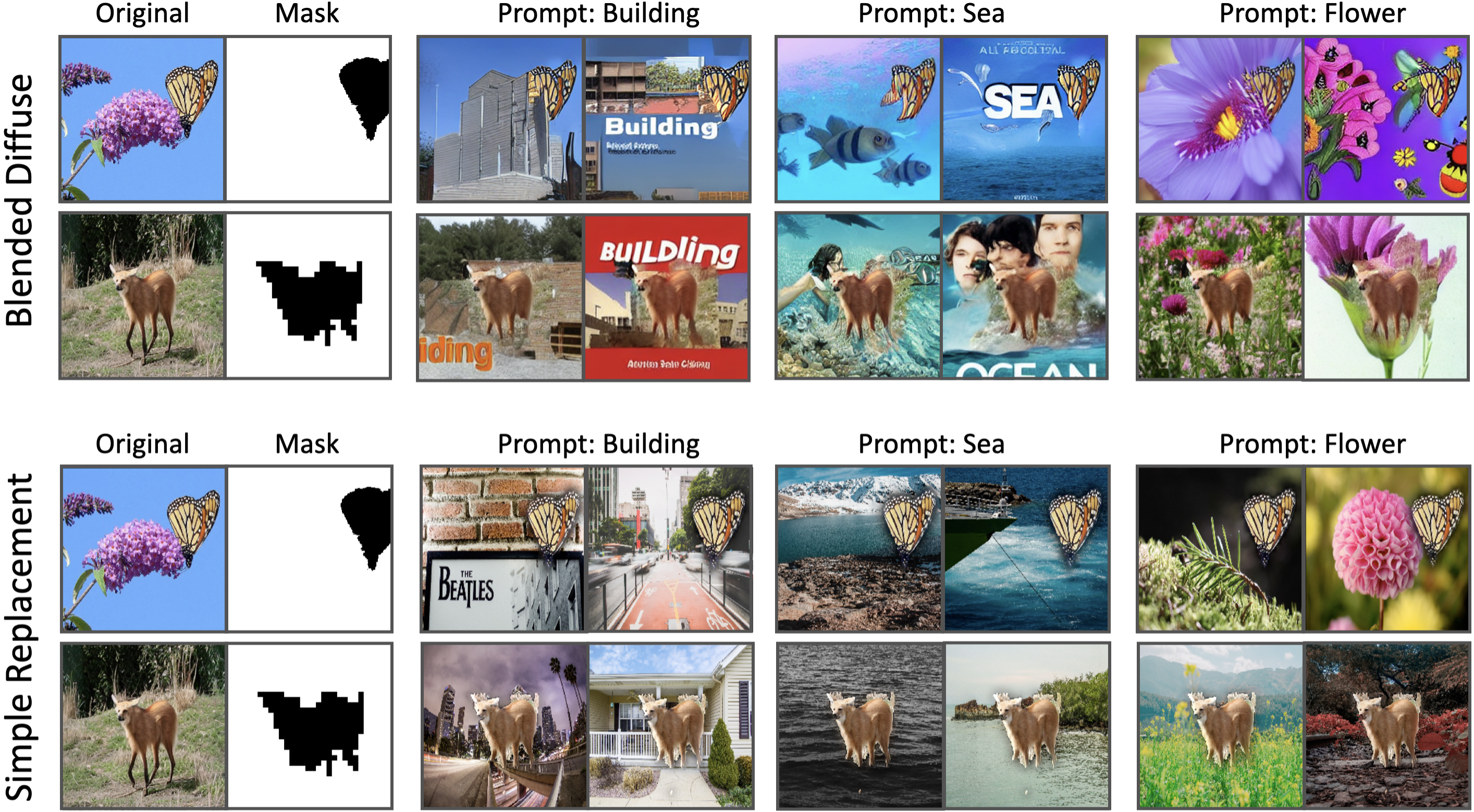}
    \caption{Background invariance testing images generated by the blended diffuse model \cite{blendedlatent} (first row), and synthesized by simple background replacement (a segmentation model \cite{ADE20k}, second row). The generated images can be more realistic, i.e., smooth blending. However, generative models can include unwanted biases, e.g., fish in the sea and insect on a flower, both of which are one of the nine foreground objects that the original models (to be tested) were trained to classify.}
    \label{fig:diffuse}
\end{figure*}

\subsubsection{Simple Background Replacement}
\label{sec:simple_bg_replace}
To test ML models that were trained to recognise or classify foreground objects, any background scenes containing any of the foreground objects should not be used. For this reason, we did not use generative algorithms to do the background replacement. We show some generated testing images using the latest generative model (blended latent diffusion \cite{blendedlatent}) in Figure \ref{fig:diffuse}. Some generated images include foreground objects due to unwanted biases, e.g., fish in the sea. This will affect the models' behaviors in an uncontrolled way. Furthermore, running large generative models can be computationally expensive, taking hours to generate one testing image. For these two reasons, we use simple background replacement Eq.\,\ref{eq:BG-Transform} together with image blending using Laplacian pyramids \cite{laplacian_blender} to remove some artifacts.

\subsection{To Analyse Testing Results} 
After we have synthesized testing images, we can 1) evaluate the diversity of the testing suite, 2) visualize the testing results to analyse the target ML model, and 3) examine if human judgements and decisions based on the visualized testing results are consistent and reliable.

\subsubsection{Diversity of Selected Testing Images}
We evaluate the diversity and comprehensiveness using the neuron coverage rate \cite{deepxplore} (percentage of triggered neurons) which has been commonly used to evaluate the extensiveness of a testing suite for ML models \cite{tensorfuzz, deephunter}. A lower neuron coverage rate often indicates monotonicity, whereas a higher coverage rate often indicates that the testing suite is sufficiently diverse, engaging a broad spectrum of the targeted ML model's internal structures \cite{survey_neuron_coverage, surprise}.

\subsubsection{Visualization of Testing Results} 

When we test an ML model $\mathbf{M}$ (trained for image classification) against the testing images, we can measure the results and intermediate results of $\mathbf{M}$ in many different locations. To reduce the number of locations (neurons) to be tested, we select the final predictions (confidence scores), and the embeddings after the final pooling layer. For background invariance testing, we can further reduce the number of neurons to be tested by utilizing the mask of the foreground objects. We feed the foreground-only image $o_i$ into $\mathbf{M}$, and obtain the top k neurons in the embedding layer. For these top $k$ neurons, we can investigate each of them or their statistics, e.g., mean or max. After we have decided the testing positions $ps$, and we can collect the response signal from $\mathbf{M}$ given a testing image $\mathbf{y}_{i,j}$, we denote the response signal as $S(\mathbf{M}, ps, \mathbf{y}_{i,j})$.

Meanwhile, we measure the \emph{semantic distance} between each testing image $\mathbf{y}_{i,j}$ and the original image $\mathbf{x}_i$ using an ensemble of ViT, CLIP, ResNet, and VGG models, which was designed for semantic image similarity \cite{our_ensemble}. 
Consider two different testing images $\mathbf{y}_{i,j}$ and $\mathbf{y}_{i,k}$ and their corresponding semantic distances to $\mathbf{x}_i$ as $d_{i,j}$ and $d_{i,k}$. The difference between their numerical measures
\begin{equation}\label{eq:Measure}
     v_{j,k} = \text{dif}(S(\mathbf{M}, ps, \mathbf{y}_{i,j}), S(\mathbf{M}, ps, \mathbf{y}_{i,k}))
\end{equation}
\noindent indicates the variation between the two testing results. As the variation corresponds to positions $d_{i,j}$ and $d_{i,k}$, this gives us a 2D data point at coordinates $p_{j,k}=(d_{i,j}, d_{i,k})$ with data value $v_{j,k}$. When we consider all the testing results for all $\mathbf{y}_{i,1}, \mathbf{y}_{i,2}, \ldots, \mathbf{y}_{i,n}$ as well as $\mathbf{x}_i$, there is point cloud with $n(n+1)$ data points in the context of $\mathbf{x}_i$.

\begin{figure}[t]
    \centering
    \includegraphics[width=\linewidth]{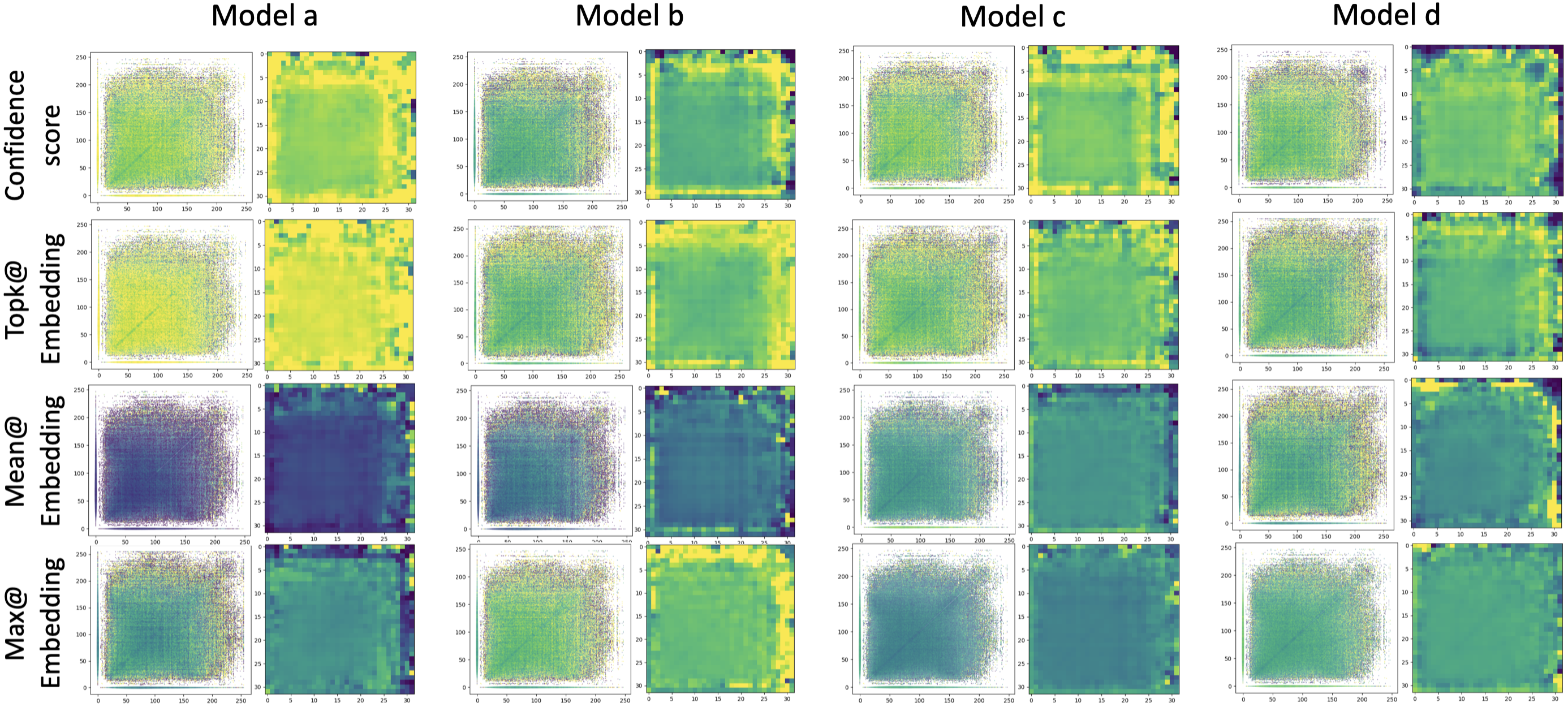}
    \caption{We show four trained ML models on IN9 dataset have different visualized testing results (of the same test run) even though they have the same accuracy score and worst-case accuracy score (averaged). We analyse these patterns in Section  \ref{sec:avg_not_good}. }
    \label{fig:bg4}
\end{figure} 

When we combine the testing results for all $l$ targeting images, we have a point cloud with $ln(n+1)$ data points, which can be visualized as scatter plots. To overcome the visualization problem of overlapping glyphs for dense areas, we adopted a common approach of radial basis functions (RBF) to transform point clouds (left column in Figure \ref{fig:bg4}) into a variance matrix (right column in Figure \ref{fig:bg4}). We detailed the adjusted RBF algorithm in Appendix D.


\subsubsection{Human Decisions and Annotations}
For one targeted ML model, based on the visualized testing results, we ask three ML practitioners to annotate if the model is: a) background invariant, b) borderline, or c) not invariant. For a model repository of 250 ML models, we evaluate the reliability and consistency of the human annotations by the inter-rater reliability \cite{irr} (IRR) between the three annotators.

\subsubsection{Comparing Different Testing Methods}
Although retrieving all similar background candidates to the original image $\mathbf{x}_i$ could lead to a high consistency (IRR) of human annotations, it might not always be desired because the selected testing suite will only include similar scenarios. 
The ideal selected testing method should include: 
\begin{itemize}
    \item diverse and comprehensive testing suites (\emph{recall}). 
    \item consistent and reliable human judgements (\emph{precision}).
\end{itemize}
In this work, we compare different testing methods, including 1) different background candidate sampling approaches (random, interval, or keyword-based), and 2) different keyword selection approaches (association ontology or vision language models, e.g., CLIP), by reporting f1 scores. We show that our method (keyword-based sampling using ontology) is the most balanced between diverse testing suites and reliable human judgements, leading to the best f1 score.

\section{Experiments}
\label{sec:exp}
%
%
To prepare ML models to be tested, we train a small repository of 250 models on the IN9 database (a smaller ImageNet with only nine classes) \cite{bgchallenge}. To guarantee that no foreground objects would appear in our testing images, we use the BG-20k \cite{bg20k} database with 20,000 background scenes as all the background candidates. We select $N=32$ background scenes for each target image, and retrieve background scenes using the keyword-based sampling or distance-based sampling methods described in Section \ref{sec:bg_sampling}. We show that similar results can be found when $N=50$ or $N=100$ in Appendix B.

For each model, we measure the signals at two positions, including the final predictions (confidence score) and the embedding layer (after the final pooling layer, or the last layer before the final MLP for Vision Transformers), as these two positions are considered the most important and interesting by our annotators. For the embedding layer, we choose a) the average value of the top 3 neurons triggered by foreground-only images, b) the maximum value, and c) the average value. This leads to one scatter plot for the confidence score and three scatter plots from the embedding layer.
For each model, we provide professional annotations based on the visualized testing results with three quality levels, namely, 1) not invariant, 2) borderline, and 3) invariant. In Appendix C, we show the statistics of the model repository, as well as a questionnaire and more details on the annotation process.

\begin{table*}[t]
\caption{Consistency, Reliability and Comprehensiveness Level of Different Background Invariance Testing Approaches}
\begin{adjustbox}{width=150mm, center}
\begin{tabular}{|l|c|cc|cc|}
\hline
\multicolumn{1}{|c|}{\multirow{2}{*}{}} & \multirow{2}{*}{\begin{tabular}[c]{@{}c@{}}Random\\ Sampling\end{tabular}} & \multicolumn{2}{c|}{Distance-based Sampling} & \multicolumn{2}{c|}{Keyword-based Sampling} \\ \cline{3-6} 
\multicolumn{1}{|c|}{}  &   & \multicolumn{1}{c|}{Nearest Top K}  & Interval (bin) & \multicolumn{1}{c|}{CLIP}  & Ontology (Ours)    \\ \hline
Neuron Coverage (recall) \cite{deepxplore} & 0.681 & \multicolumn{1}{c|}{0.133}  & 0.667  & \multicolumn{1}{c|}{0.591}    & 0.652   \\ \hline
Fleiss’ reliability (precision) \cite{fleiss1971measuring}  & 0.384  & \multicolumn{1}{c|}{0.906}  & 0.531 & \multicolumn{1}{c|}{0.640}    & 0.649   \\ \hline
F1 score   & 0.491 & \multicolumn{1}{c|}{0.232}  & 0.591  & \multicolumn{1}{c|}{0.615}  & \textbf{0.650}       \\ \hline
\end{tabular}
\end{adjustbox}
\label{table:irr}
\end{table*}

\subsection{Worst-case Accuracy versus Visual Analysis}
\label{sec:avg_not_good}
We first confirm that visualization-based testing is more informative than global statistics. We show that the visualized performance (variance matrices) of four models can be very different even though they have the same (worst-case) accuracy score in Figure \ref{fig:bg4}. The patterns of $\mathbf{M}_a$ suggest that $\mathbf{M}_a$ gives wrongly predicted testing images high confidence scores, and the outputs of the top-3 neurons triggered by foreground-only images are not consistent, which indicates that $\mathbf{M}_a$ might rely more on the background to make its decisions instead of the foreground objects. The abrupt green pattern at the top-left for $\mathbf{M}_c$ might be a signal of data leakage which needs further examination before being deployed. The yellow pattern on the edges of $\mathbf{M}_b$ suggests that $\mathbf{M}_b$ treats a few of testing images differently from the others. This could suggest that $\mathbf{M}_b$ might be sensitive to the presence of certain objects. For these reasons, our three annotators labelled $\mathbf{M}_a$ and $\mathbf{M}_c$ as \emph{failed}, $\mathbf{M}_b$ as \emph{borderline}, and $\mathbf{M}_d$ as \emph{passed}. These four models have the same statistical evaluation scores (worst-case accuracy) whilst their visualized testing results can yield different judgements on their invariance qualities. To consolidate if we can use such visualization-based testing framework for background invariance testing, we evaluate 1) how diverse the testing suites are, and 2) how consistent the human annotations are.

\subsection{Comparisons Between Testing Methods}
Human judgements of invariance qualities based on visualized testing results might vary across different testing runs. Therefore, we evaluate whether a testing method is desired based on: 1) diversity -- neuron coverage rate (\emph{recall}), and 2) consistency -- inter-rater reliability (IRR) score (\emph{precision}) as discussed in Section 4.3. Neuron coverage rates are often used to evaluate whether a selected testing suite is diverse enough to cover as many scenarios as possible \cite{survey_neuron_coverage, deepxplore}, and IRR scores are often used to evaluate the consistency and reliability of professional annotations \cite{irr} between different practitioners. We report f1 score (between diversity -- recall, and reliability -- precision) when comparing different testing methods.

\subsubsection{Prior Works: Random/Nearest Sampling} In Table \ref{table:irr}, we show that although the random sampling testing approach resulted in the best neuron coverage rate, the IRR score is relatively lower because of the randomness. Meanwhile, selecting the nearest top-$k$ background scenes resulted in the highest inter-rate reliability score, however, the neuron coverage rate (diversity level) is the lowest. Therefore, these two testing methods might not be the most suitable for background invariance testing.

\subsubsection{Distance and Keyword-based Sampling} As shown in Table \ref{table:irr}, although distance-based sampling (interval) can achieve a higher neuron coverage rate, the resultant human judgements are not most consistent, whereas keyword-based sampling methods are the most balanced between diversity and reliability. Among keyword-based sampling methods, CLIP \cite{clip} tends to find similar (matched) items for the target image, which leads to less diverse testing suites. Meanwhile, our ontology expands the originally detected keywords using association analysis and thus becomes the most balanced between diversity and consistency (i.e., the best f1 score). We also show the distribution of the distances between target images and the testing images synthesized using random, nearest neighbor, and our ontology-based method in Figure \ref{fig:dist_distribution} in Appendix B5.


\begin{table}[t]
\centering
\caption{Automation results: the automation accuracy using random forest is around 80\% and the inter-rater reliability score with majority votes is around 0.65.}
\begin{adjustbox}{width=80mm, center}
\begin{tabular}{|c|c|c|}
\hline
 & Automation Accuracy & IRR Score   \\ \hline
Random Forest  & 79.7 $\pm$ 7.5\%        & 0.649 $\pm$ 0.091 \\ \hline
AdaBoost       & 74.8 $\pm$ 9.1\%        & 0.599 $\pm$ 0.102 \\ \hline
\end{tabular}
\end{adjustbox}
\label{table:results}
\end{table}

\subsection{Automated Background Invariance Testing}
To avoid the labor-intensive manual analysis process, we investigate if the entire testing procedure can be automated. We split the model repository into a training set (2/3 of the models) and a testing set (1/3 of the models). And we train a simple random forest to judge if models are rated as passed, failed or borderline using some hand-crafted features from the variance matrices \cite{sparselayers}. To make the results more statistically significant, we randomly split the data, repeated the experiments ten times and reported the averaged results and the standard deviation. In Table \ref{table:results}, we show that we can achieve around 80\% automation accuracy. Furthermore, the IRR scores between the predictions from assessors and the majority votes are similar to those of the three coders ($\sim$0.65). Therefore it shows the proposed framework can work as a fully automated background testing mechanism with sufficient accuracy.

\section{Conclusion}
In this work, we first confirm that visualization-based testing methods are more informative than reporting global statistics for background invariance testing. We show that models having the same averaged accuracy score can perform differently. With the proposed framework, we can visualize the testing results and find some visual patterns which facilitate further analysis of the invariance qualities.

We identify the challenges of utilizing visualization-based testing techniques when the data space is gigantic and adequate sampling is not feasible. We find that randomly sampled testing examples leads to inconsistent visualized patterns, hence inconsistent human decisions on invariance qualities across different testing runs. Meanwhile, nearest-neighbor sampling leads to consistent human judgment but with limited diversity in the sampled testing examples. 

We propose an association ontology-based approach that can lead to 1) diverse testing suite, and 2) consistent and reliable human judgments on the invariance qualities of interested ML models. In the future, a larger model repository can be collected to confirm the feasibility of the framework.



{
    \small
    \bibliographystyle{ieeenat_fullname}
    \bibliography{main}
}
\newpage
\appendix

\noindent \textbf{Appendix}

\section{Building the Ontology}

\begin{figure}[h]
    \centering
    \includegraphics[height=40mm]{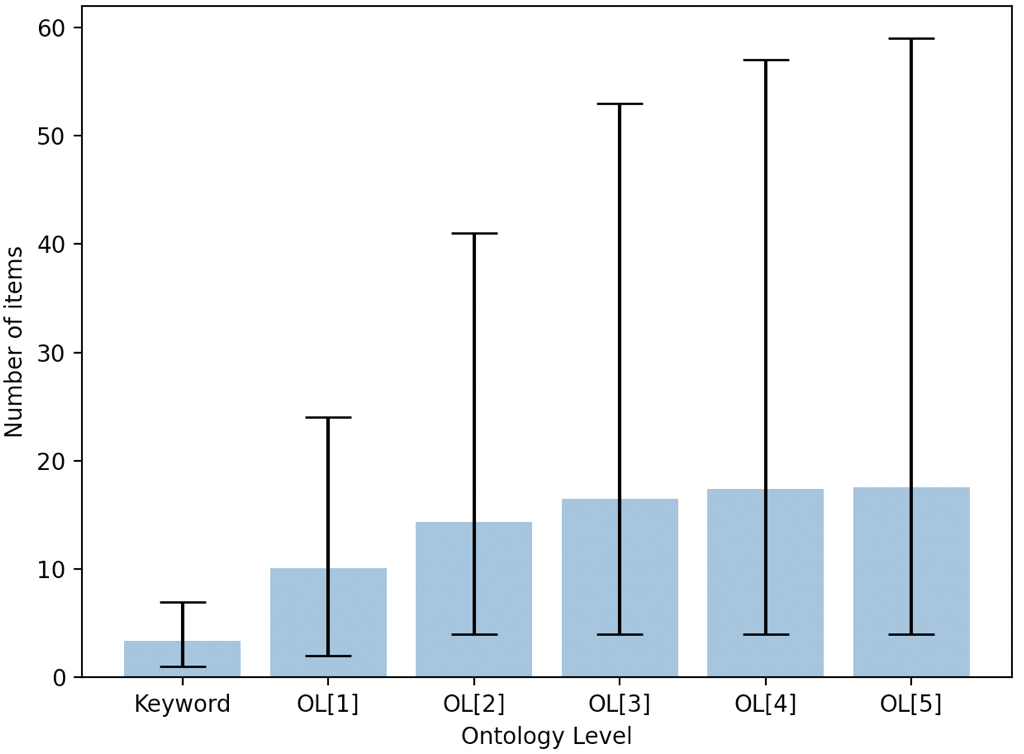}
    \caption{Number (maximum, mean, minimum) of enriched / expanded items after each ontology level.}
    \label{fig:nbitems}
\end{figure}

\paragraph{A1. Semantic Keyword Vectors}
For any given input image $\mathbf{x}$, we have a keyword vector $\vec{v}(\mathbf{x})=[v_1, v_2, \ldots]$ (the value $v_i$ is corresponding to the $i^{th}$ keyword) obtained by pre-trained scene understanding models. In this work, we use $k=150+365=515$ keywords (excluding foreground objects). We define the keyword vector in the following format:
\begin{itemize}
    \item $\vec{v}_{seg}$ $\rightarrow$ $v_{1}$ to $v_{150}$: object predictions, e.g., wall, floor, etc. These 150 object predictions are provided by a pre-trained segmentation model on the ADE20k dataset.
    \item $\vec{v}_{scene}$ $\rightarrow$ $v_{151}$ to $v_{525}$: scene class predictions, e.g., pond, grassland, etc. These 365 scene class predictions are provided by a pre-trained model on Place365 dataset.
\end{itemize}
We define the $k^{th}$ keyword as present if $v_k > 1\%$ when the $k^{th}$ keyword is a segmentation keyword, or if $v_k = \max(\vec{v}_{scene})$ when the $k^{th}$ keyword is a scene classification keyword. In Figure \ref{fig:nbitems}, we show that on average, one input image has three frequent items. However, this is not enough for background invariance testing, therefore, we build an ontology to expand the keywords.

\paragraph{A2. Databases} \textbf{For scene understanding}: we use pre-trained models on two databases. In the ADE20k dataset \cite{ADE20k}, 150 object classes in 20,000 background scenes from the SUN and Places database are annotated. Some labelled objects can be a whole object or a part of another object, e.g., a door can be an indoor picture or a part of a car. Place365 dataset consists of 1,803,460 training images with 365 different scene classes \cite{place365}. And the number of images per class varies from 3,068 to 5,000. \textbf{For preparing models} to be tested: we use a smaller version of ImageNet, namely IN9 which consists of images from ImageNet but labelled with nine classes \cite{bgchallenge}. There are 45,405 images on the training set (5045 images per class) and 4185 images on the testing set (465 images per class). They also provide segmented foreground masks for some images on the testing set. Finally, \textbf{For background candidates}: we use BG20k which consists of 20,000 background scenes with no foreground images \cite{bg20k}.

\paragraph{A3. Association Analysis and Ontology:}
We use association analysis to build an ontology to help background enrichment in this work. Ontology is a graph based approach to store relationships between different entities, e.g., keywords in our case. Commonly used  relationships in an ontology include ``$\alpha$ is related to $\beta$" or ``$\alpha$ has $\beta$". In this work, the relationship in our ontology is defined:
``Keyword A is related to keyword B in a manner
that if A is known to be in an image, B has the conditional probability of $pr(B|A)$ to occur in the same image.
This probability is computed as the confidence in association analysis (Eq.\,\ref{eq:Confidence}).

\paragraph{A4. Pre-steps of Building the Ontology}
\begin{itemize}
    \item Run a pre-trained scene understanding model on each of the given input images and generate a keyword vector for the image.
    \item Run association analysis to obtain frequent items, itemsets, association rules, and the ontology using the Apriori or FP-Growth algorithm \cite{fptree}.
    \item The ontology is re-built only when necessary, e.g., when new keywords are added into the ontology, or when the distribution of the dataset has changed.
\end{itemize}

\paragraph{A5. Keyword Expansion}
The major steps are:
\begin{enumerate}
    \item For an input image, obtain a keywords vector $V_0$ using the scene understanding model. Each keyword is marked with [0] indicating level 0.
    Initialize the overall keywords vector $V$ as $V \leftarrow V_0$, and $i \leftarrow 0$
    \item If the overall keywords vector $V$ has reached the minimal number of keywords desired, i.e., $\|V\| \geq \text{MINKWS}$
    or the number of iterations reached the maximum number allowed, i.e., $i \geq \text{MAXITS}$,
    stop the process and return the overall keywords vector $V$. Otherwise go to the next step.
    \item We use the ontology to find all keywords that connected to the keywords in the vector $V_{i}$. For any newly-found keyword that is not already in $V$, add it to $V_{i+1}$, and mark it as $[i+1]$, where $i+1$ indicates the new extended level.
    \item $V \leftarrow \text{union}(V_0, V_1, \ldots, V_{i+1})$, $i \leftarrow i+1$.
    
\end{enumerate}


    



We use the keyword expansion algorithm to enrich the originally detected keywords $V_0$ (from the pre-trained scene understanding model). In Figure \ref{fig:nbitems}, we show that using the ontology with more iterations can increase the number of total keywords in $V$. However, due to the limited size of the ontology (the limited size of the database we use to build the ontology), the effect of using the ontology becomes less significant after the 4th iteration, which is consistent with the experiments in Section \ref{sec:exp}.


\begin{table*}[t]
\centering
\caption{Experiment results: the automation accuracy using random forest as the assessor is around 80\% and the inter-rater reliability score with majority votes is around 0.65. Meanwhile, no significant differences are found using different settings (i.e., kernel size, sigma value) for RBF interpolation.}
\begin{adjustbox}{width=\textwidth}
\begin{tabular}{|c|ccc|ccc|}
\hline
\multirow{2}{*}{\diagbox[width=33mm]{Assessor}{Interpolator}}   & \multicolumn{3}{c|}{Automation Accuracy} & \multicolumn{3}{c|}{Inter-rater Reliability with Majority Votes} \\ \cline{2-7} & \multicolumn{1}{c|}{Size: 32, $\sigma$=2} & \multicolumn{1}{c|}{Size: 16, $\sigma$=5} & Size: 32, $\sigma$=10 & \multicolumn{1}{c|}{Size: 32, $\sigma$=2} & \multicolumn{1}{c|}{Size: 16, $\sigma$=5} & Size: 32, $\sigma$=10 \\ \hline
Random Forest       & \multicolumn{1}{c|}{79.6$\pm$8.1\%}  & \multicolumn{1}{c|}{78.7$\pm$7.7\%}  & \textbf{79.7$\pm$7.5\%}   & \multicolumn{1}{c|}{\textbf{0.651$\pm$0.103}}   & \multicolumn{1}{c|}{0.635$\pm$0.116}   & 0.649$\pm$0.091    \\ \hline
Adaboost            & \multicolumn{1}{c|}{54.5$\pm$7.5\%}  & \multicolumn{1}{c|}{70.9$\pm$10.2\%}  & 74.8$\pm$9.1\%   & \multicolumn{1}{c|}{0.358$\pm$0.134}   & \multicolumn{1}{c|}{0.548$\pm$0.159}   & 0.599$\pm$0.102    \\ \hline
Worst-case accuracy & \multicolumn{3}{c|}{64.4\%} & \multicolumn{3}{c|}{0.387} \\ \hline
\end{tabular}
\end{adjustbox}
\label{table:results_rbfs}
\end{table*}


\section{Ablation Study}

\paragraph{B1. Number of Selected Testing Images}
When use the two sampling approaches (see Section \ref{sec:bg_sampling}) to search testing scenes, we obtain $N=32$ testing images via 32 intervals (bins) or 32 keywords. Here we show that we obtain similar results when $N=50$ or $N=100$.

As shown in Table \ref{table:ablation_f1}, selecting different number of testing images leads to similar neuron coverage rates and IRR scores. This is aligned with \cite{deepxplore} where fuzzing techniques (e.g., mutation) are used to find adversarial examples to achieve a better coverage rate. Therefore in this work, we choose $N=32$.

\begin{table}[h]
\centering
\caption{Selecting different number of testing images leads to similar Recall, precision and F1 score}
\begin{adjustbox}{width=\linewidth}
\begin{tabular}{|l|c|c|c|}
\hline
\multicolumn{1}{|c|}{}               & N=32  & N=50  & N=100 \\ \hline
Neuron Coverage (Recall)             & 0.652 & 0.657 & 0.660 \\ \hline
Fleiss' Reliability (IRR, Precision) & 0.649 & 0.645 & 0.645 \\ \hline
F1                                   & 0.650 & 0.651 & 0.652 \\ \hline
\end{tabular}
\end{adjustbox}
\label{table:ablation_f1}
\end{table}

\begin{figure}[t]
    \centering
    \includegraphics[height=87mm]{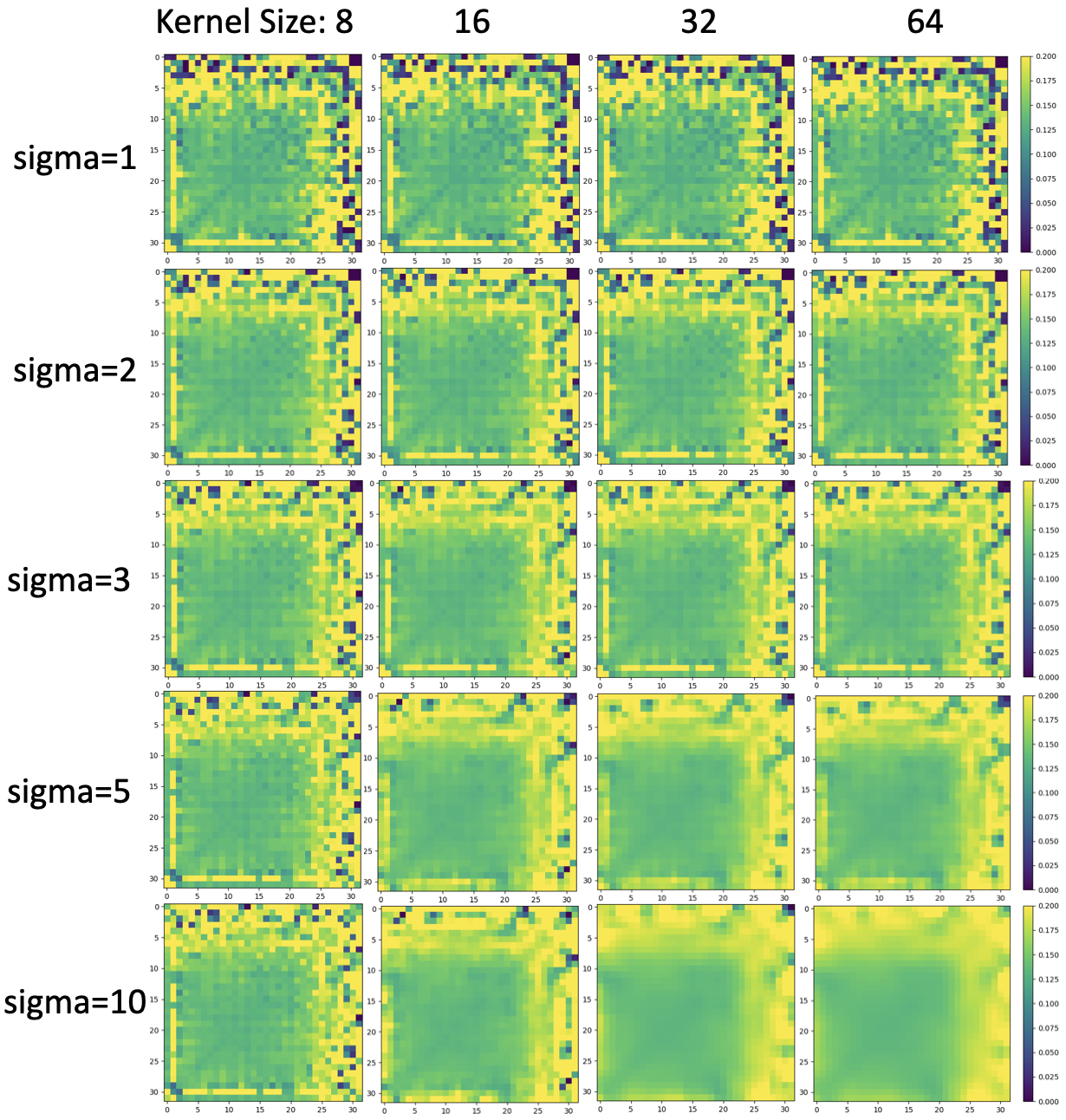}
    \caption{Different interpolant parameters for  $\mathrm{M_{61}}$. We use (Kernel size 16, sigma 5), (Kernel size 32, sigma 2), and (Kernel size 32, sigma 10) respectively in Figure \ref{fig:percent}.}
    \label{fig:rbf_ablation}
\end{figure}

\begin{figure}[t]
    \centering
    \includegraphics[width=\linewidth]{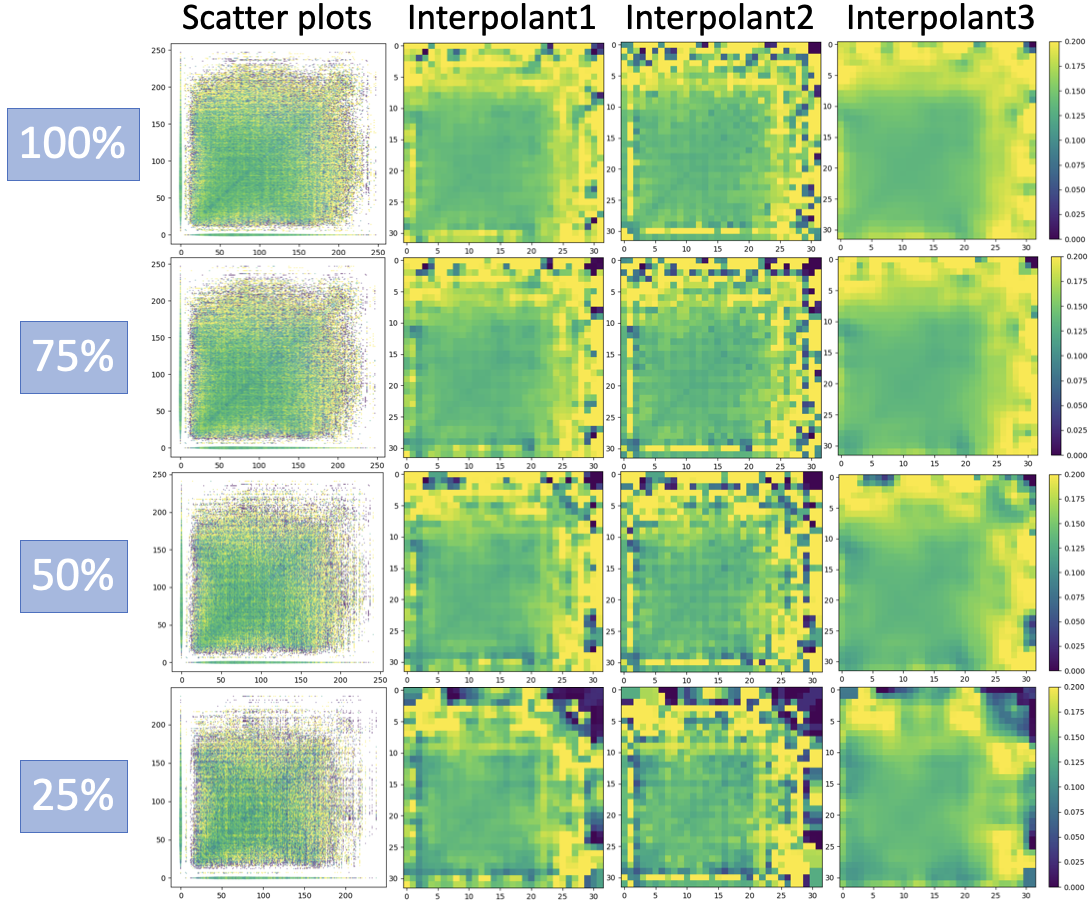}
    \caption{We use different numbers (percentages) of the testing images to generate scatter plots. The number of images will not significantly affect the quality of the generated plots unless the number becomes too small, e.g., $\leq$ 25\%. The example we show here is model $\mathrm{M_{61}}$}
    \label{fig:percent}
\end{figure}

\paragraph{B2. Different Settings for RBF Interpolation}
We conduct the ablation study for some parameters of the interpolation, i.e., the kernel size and sigma values. In Figure \ref{fig:percent}, the three interpolants are (kernel size: 16, sigma: 5, $K$: 32), (kernel size: 32, sigma: 2, $K$: 32) and (kernel size: 32, sigma: 10, $K$: 32) respectively. The $K$ values only affect a few points at the corners and we barely see any difference. As shown in Figure \ref{fig:rbf_ablation}, kernel size and sigma value affect how blurry the interpolated scatter plots are. Based on the quality of the interpolated scatter plots, we tested these three interpolants and reported the automation results in table \ref{table:results_rbfs} where we show that using different parameters for the interpolation did not heavily affect the automation accuracy (the performance of assessors). 

\paragraph{B3. Number of Testing Images} To generate the scatter plots for a given model, we use the testing set of IN9 dataset \cite{bgchallenge}. However, among all the 4185 images in the testing set, only 1712 images are provided with a foreground mask. Therefore, we use the 1712 images to generate the scatter plots. In Figure \ref{fig:percent}, we show that using different percentages of the 1712 images will not significantly affect the qualities of the generated scatter plots (neither the original nor interpolated). However, when the number of images becomes too low, e.g., $\leq$25\% (less than 430 images), the generated scatter plots are starting to be heavily affected.



\begin{figure}
    \centering
    \includegraphics[width=\linewidth]{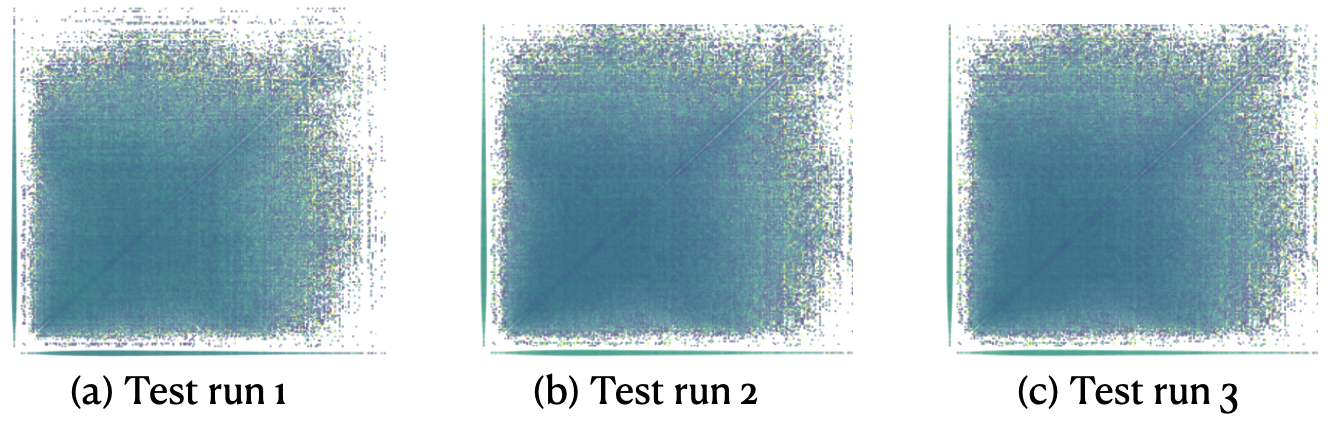}
    \caption{Our proposed methods lead to consistent visualized testing patterns across multiple different testing runs, compared with random sampling (Figure \ref{fig:VarianceMatrix})}
    \label{fig:ontology3run}
\end{figure}

\paragraph{B4. Testing Patterns Across Different Testing Runs} Finally, we show that our proposed testing method leads to consistent visualized patterns across different runs (in Figure \ref{fig:ontology3run}), especially when compared with random sampling (Figure \ref{fig:VarianceMatrix}). This is also reflected by the relatively higher reliability across annotators.

\begin{figure}[t]
    \centering
    \includegraphics[width=\linewidth]{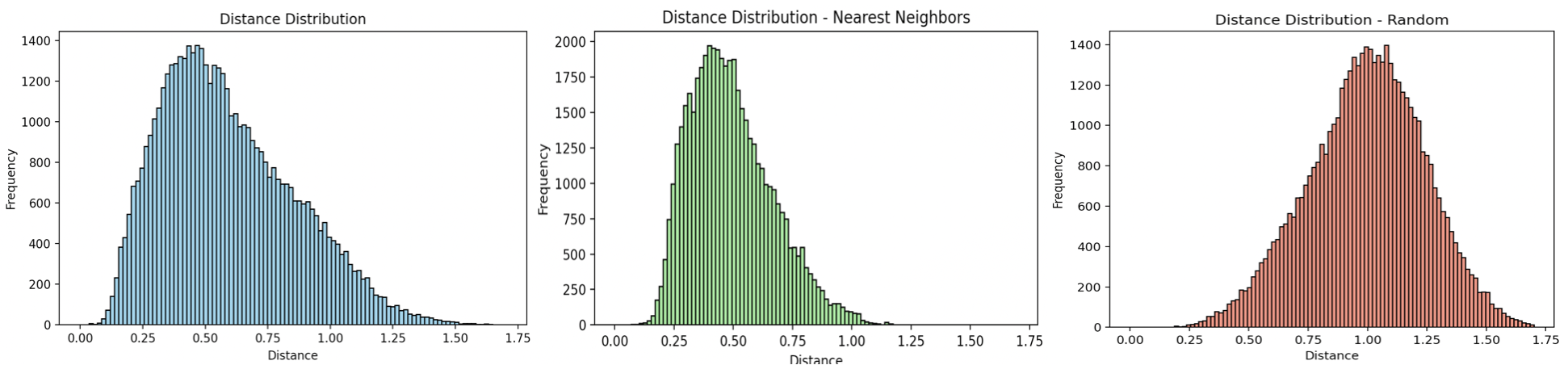}
    \caption{Distribution of distances between original images and synthesized testing images. Left (blue): Association Ontology, middle (green): nearest neighbors, right (red): random.}
    \label{fig:dist_distribution}
\end{figure}

\paragraph{B5. Distribution of the distances $d_{i, j}$} As shown in Figure \ref{fig:dist_distribution}, the distances are Gaussian distributed. As expected, compared with nearest neighbor, the distribution (of association ontology) suggests that our sampled scenes are more diverse. Compared with random sampling, the distribution shows that the sampled scenes prioritize scenes with a stronger association with the original images. Any methods involving selecting $N$ items from $K (\gg N)$ has a long-tail by definition. Only when the sorting of $N$ is not good, it becomes a problem. In this work, we sample testing scenes based on semantically connected keywords. In our approach, the ``semantic connections'' depend on the frequency of association.


\section{Automated Background Invariance Testing}

We build a small model repository of 250 models for object classification. 
The models were trained under different settings:
\begin{itemize}
    \item Architectures: VGG13bn, VGG13, VGG11bn, VGG11 \cite{vggnet}, ResNet18 \cite{resnet}, and Vision Transformer \cite{vit}
    \item Hyper-parameters: learning rate, batch size, epochs
    \item Augmentation: rotation, brightness, scaling, using images with only foreground (black pixels as background)
    \item Optimizers: SGD, Adam, RMSprop
    \item Loss: cross-entropy loss, triplet loss, adversarial loss
\end{itemize}

We apply these models to the synthesized testing images.
With the two measuring positions per model, we generate four variance matrices using the results from the ML Testing process as mentioned in Section \ref{sec:exp}. In this section, we discuss our professional annotations based on the variance matrices, statistics of our model repository, and finally analysis on the automation process of the invariance testing procedure.

\paragraph{C1. Professional Annotations}
We survey the three professionals who provide the annotations for background invariance testing. We ask them the following questions:
\begin{itemize}
    \item Q1: Which locations are you interested in?
    \item Q2: Are the interpolated scatter plots helpful? Which interpolant is the most helpful?
    \item Q3: Are there any common visual patterns?
    \item Q4: Are the annotations aligned with the worst-case accuracy?
    \item Q5: Would you say your decisions are consistent? Are you confident about your decisions?
    \item Q6: Any interesting findings about the models?
\end{itemize}

\begin{figure}[t]
    \centering
    \includegraphics[height=90mm]{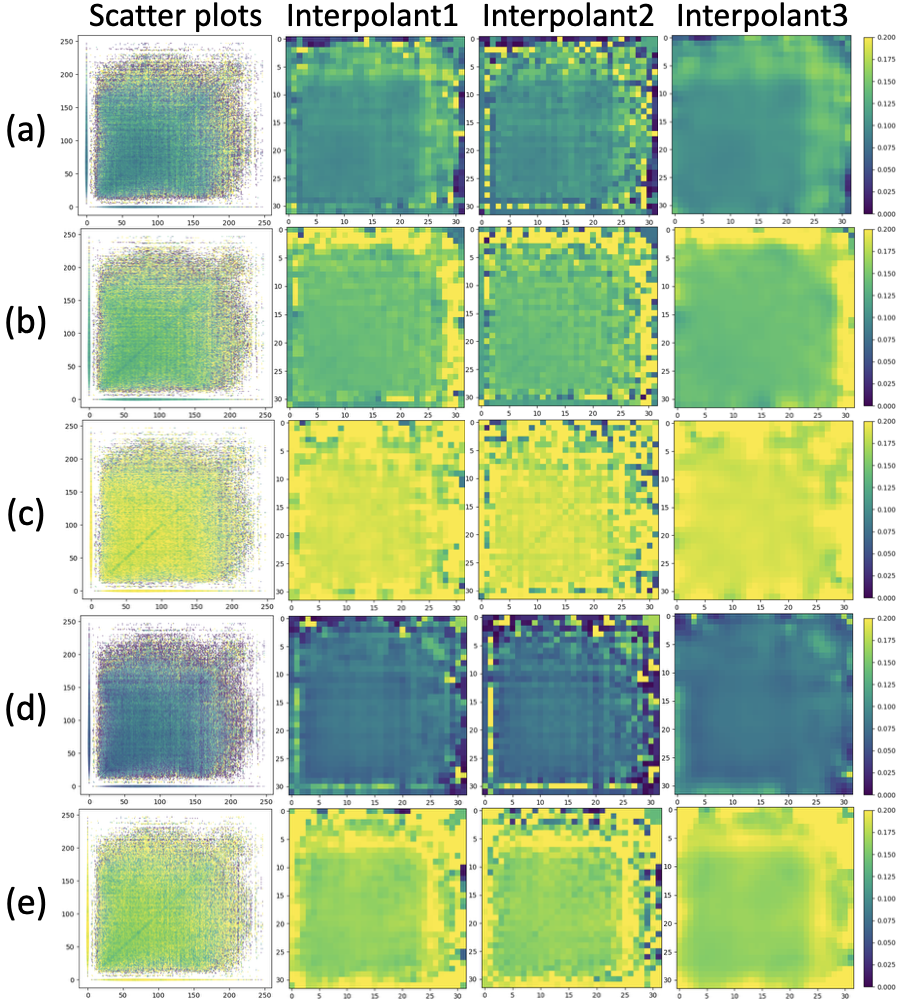}
    \caption{Examples of original scatter plots and interpolated / accumulated scatter plots. From (a) to (e) are model $\mathrm{M_{1}}$, $\mathrm{M_{7}}$, $\mathrm{M_{16}}$, $\mathrm{M_{68}}$, $\mathrm{M_{71}}$ respectively.}
    \label{fig:rbf_many}
\end{figure}

\paragraph{Q1: Testing locations} All the three coders agree that the final predictions and outputs from the last layer before the final fully connected are important to be tested.

\paragraph{Q2: Helpful interpolated} Coder 1 and 2 mention all of the interpolated plots are helpful and there is no particular interpolator that is superior to the others. Coder 3 mentions they check all of the interpolated plots and find the one with kernel size 32 and $\sigma=10$ helps their judgements the most because it shows the most straightforward global shape of the plots.

\paragraph{Q3: Common visual patterns} Coder 1 and 3 mention that the most common patterns are those with a green area (which indicates a lower error rate) at the bottom left, and some yellowish patterns (which indicates a higher error rate) at either the edges or corners at the top right. Coder 2 notices it is common that there is a green line along the primary diagonals (from bottom left to top right) of the plots. We show more different patterns in Figure \ref{fig:rbf_many}.

\paragraph{Q4: Annotations versus worst-case accuracy} All three coders mention their annotations are not completely aligned with worst-case accuracy. However, they all find their annotations and the traditional metric are related to each other to some extent.

\paragraph{Q5: Consistent annotations} All three coders are confident about their decisions for the majority of the models. However, they are less confident for those labelled as ``borderline".

\begin{figure}[t]
    \centering
    \includegraphics[width=\linewidth]{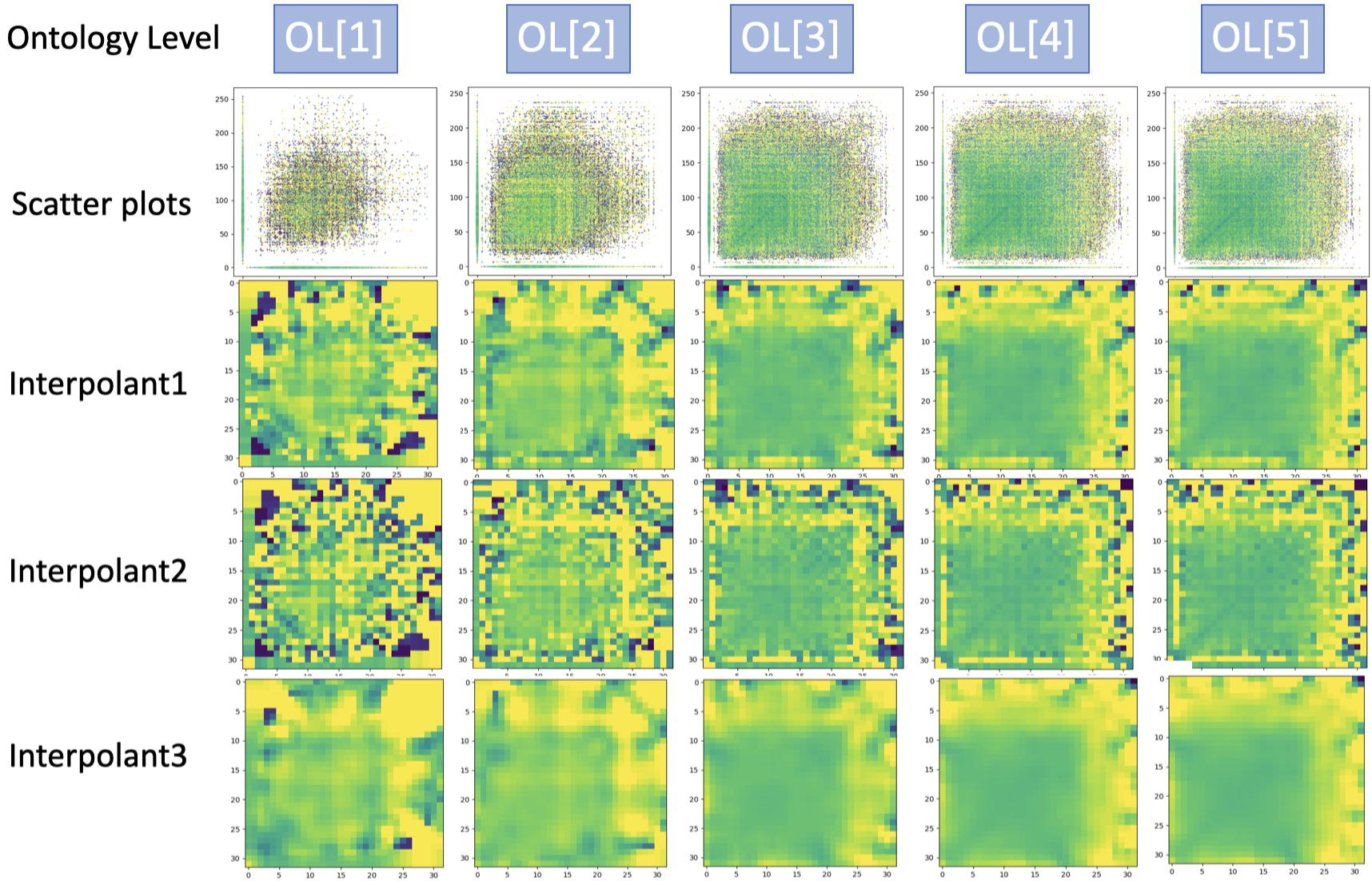}
    \caption{Original scatter plots (first column) and interpolated variance matrices (second to fourth column). The scatter plots are generated using the ontology level one to five, i.e., $OL[1], OL[2], $ $\ldots, OL[5]$ for a randomly selected model $\mathrm{M_{61}}$. Interpolant 1 - 3: (kernel size 16 $\sigma$=5), (kernel size 32 $\sigma$=2), and (kernel size 32 $\sigma$=10).}
    \label{fig:SP2RBF}
\end{figure}

\paragraph{Q6: Interesting findings} Coder 1: vision transformers seem to be naturally more robust against background permutations. Even when the final predictions do not appear to be robust at all, their last layer before the head module can still be robust (dark green). Coder 2: models trained using images without background (black pixels as their background) appear to be more robust than those trained with natural images. Coder 3: metric learning, e.g., triplet loss, does not seem to be helpful even for the convolutional layers that are used to form the triplet loss function.

\begin{figure}[t]
    \centering
    \includegraphics[height=27mm]{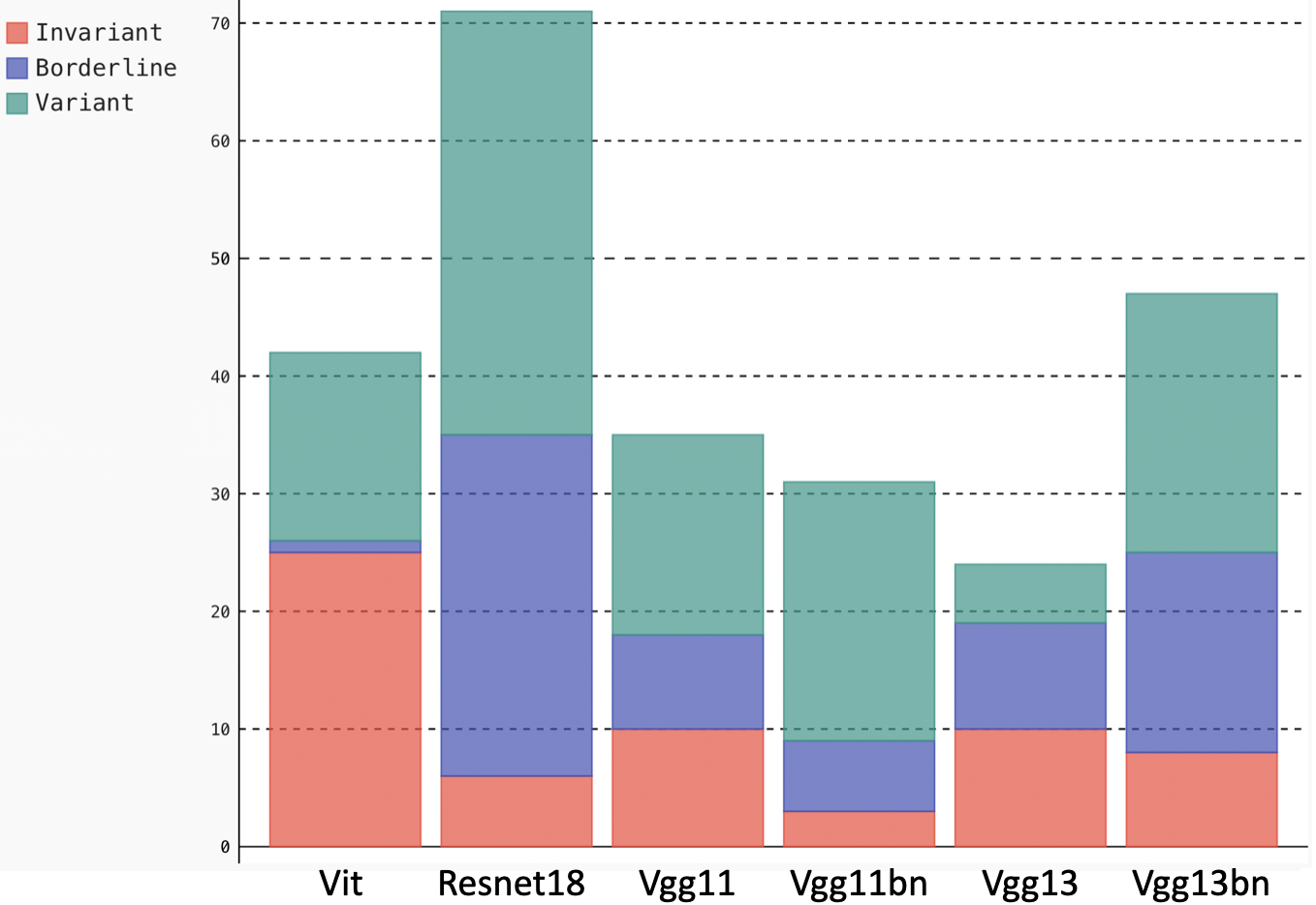}
    \includegraphics[height=27mm]{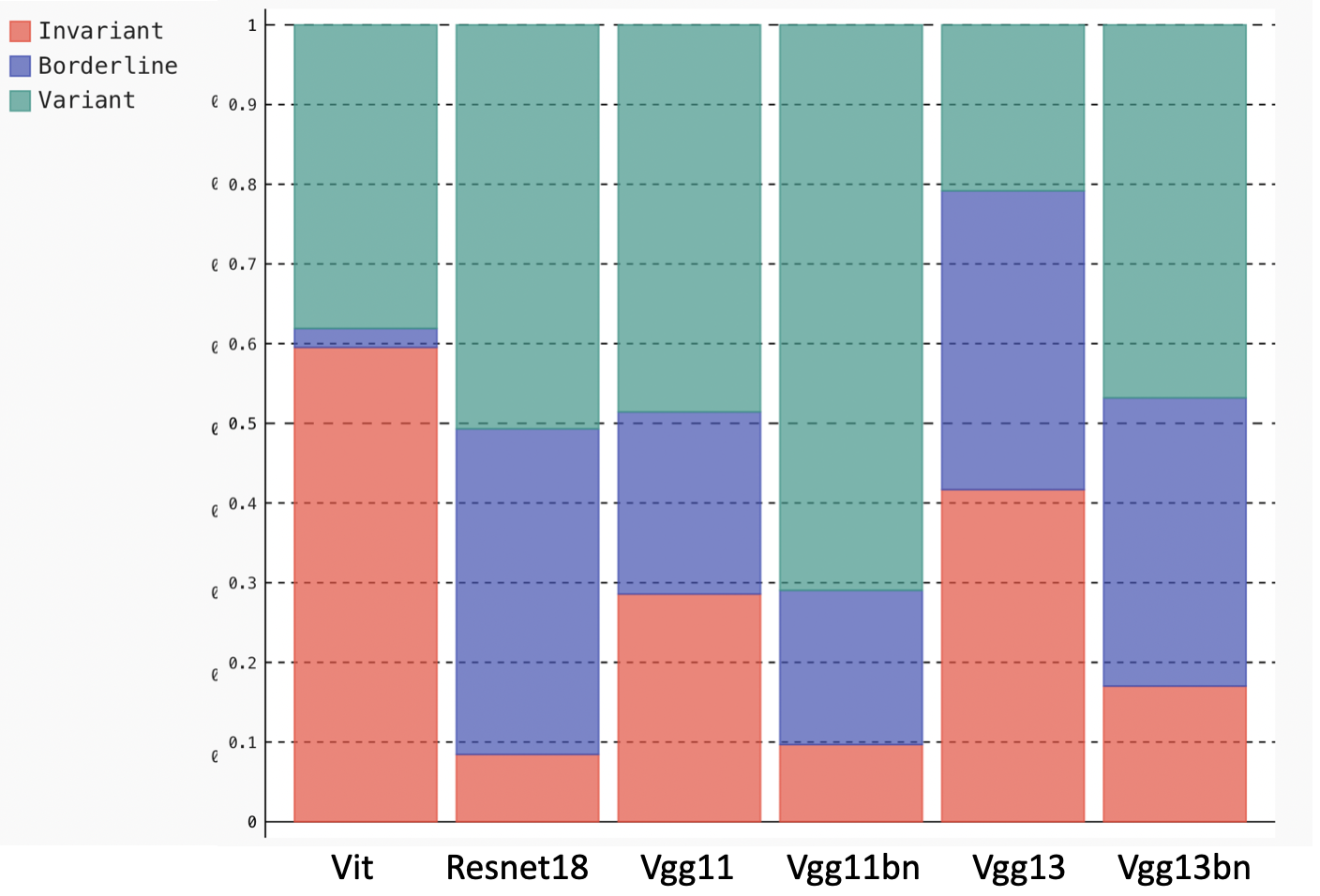}
    \caption{Statistics of model architectures in our model repository. Left: number of models with each architecture in our model repository. Right: percentage of invariant / borderline / variant models for each architecture in our model repository.}
    \label{fig:arch_stat}
\end{figure}

\begin{figure}[t]
    \centering
    \includegraphics[width=\linewidth]{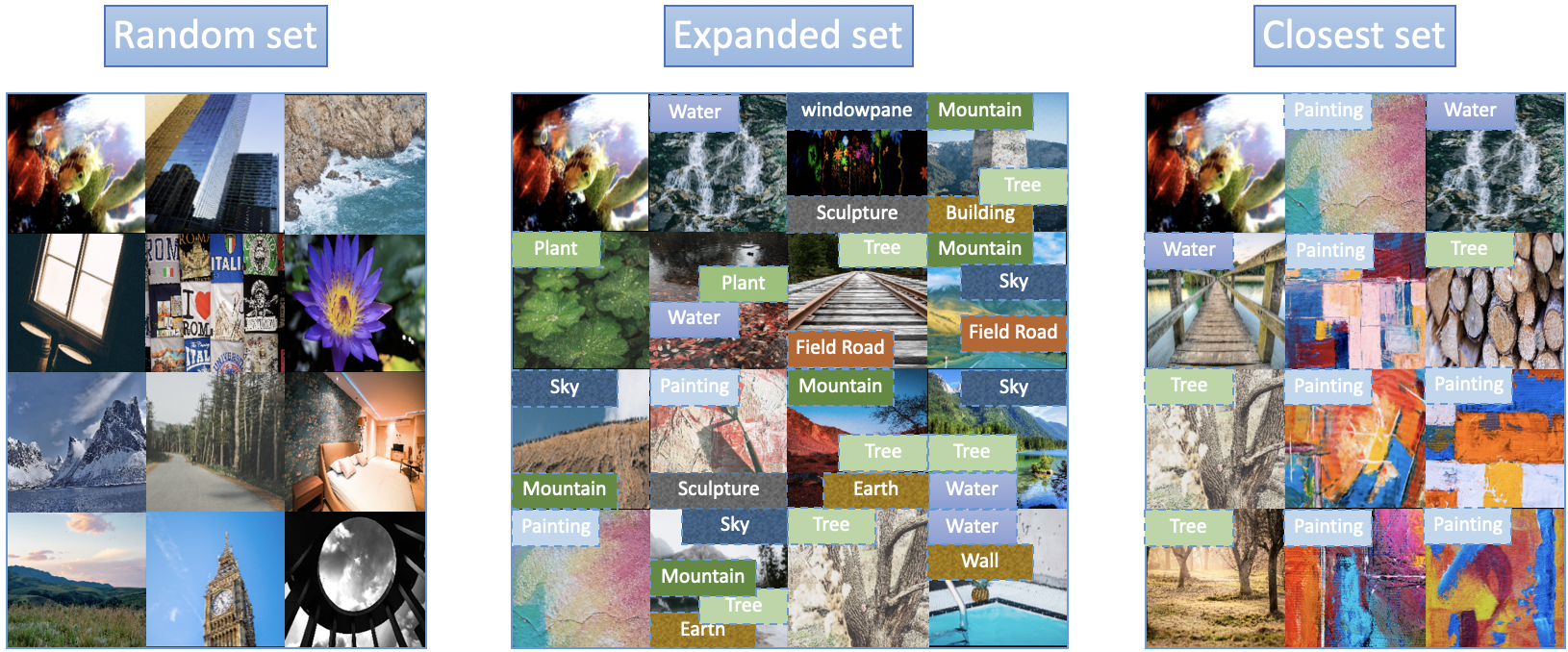}
    \caption{Three sets of example background scenes discovered for a target image of fish (the top-left of each set). The random set includes mostly unsuitable images. The closest set includes those discovered using only the original keywords $K_x = \{$painting, water, tree$\}$. The expanded set includes those discovered using the ontology, showing more suitable background scenes. Note that those keywords found in background scenes should not include any of the foreground objects that the original ML models were trained to classify as specified at the beginning of Section \ref{sec:mehod}.
    }
    \label{fig:Similarity}
\end{figure}  
\paragraph{C2. Model Repository}

In this section, we show some statistics of our model repository (250 models). As mentioned in the main paper, we trained the models under different settings, and provide professional annotations for each of them as being background-invariant, borderline or background-variant.

\begin{table}[h]
\centering
\caption{Statistics of the annotations of the model repository}
\begin{tabular}{|c|c|c|c|}
\hline
 Invariant   & Borderline   & Variant      & Total       \\ \hline
 62 (24.8\%) & 70 (28.0\%) & 118 (47.2\%) & 250 (100\%) \\ \hline
\end{tabular}
\label{table:stat_annotation}
\end{table}

As shown in Table \ref{table:stat_annotation}, the model repository is not balanced. There are more background variant models than background invariant models, which is expected as most of the models were trained using no special technique to boost their background invariance qualities; Unlike other types of transformations, e.g., rotation, background augmentation is not commonly used.

\begin{table}[h]
\centering
\caption{Statistics of the accuracy of the model repository}
\begin{tabular}{|c|c|c|c|c|}
\hline
               & mean    & max     & min     & std     \\ \hline
Accuracy       & 91.88\% & 99.42\% & 43.40\% & 9.11\%  \\ \hline
Worst-case & 28.76\% & 65.36\% & 0.18\%  & 11.92\% \\ \hline
\end{tabular}
\label{table: stat_acc}
\end{table}

\begin{figure*}[t]

    \centering
    \includegraphics[width=\textwidth]{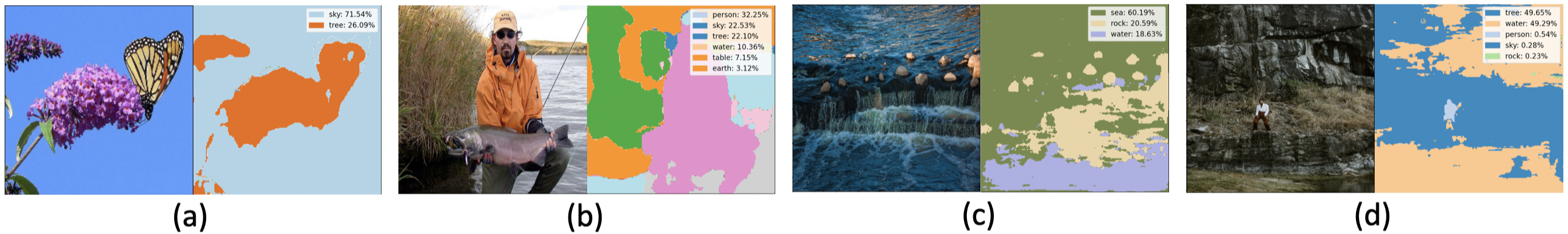}
    \caption{(a), (b) are target images. (c), (d) are background candidates. (a) and (c) have only a few detected keywords.}
    \label{fig:scene_item}
    
    \centering
    \includegraphics[width=\textwidth]{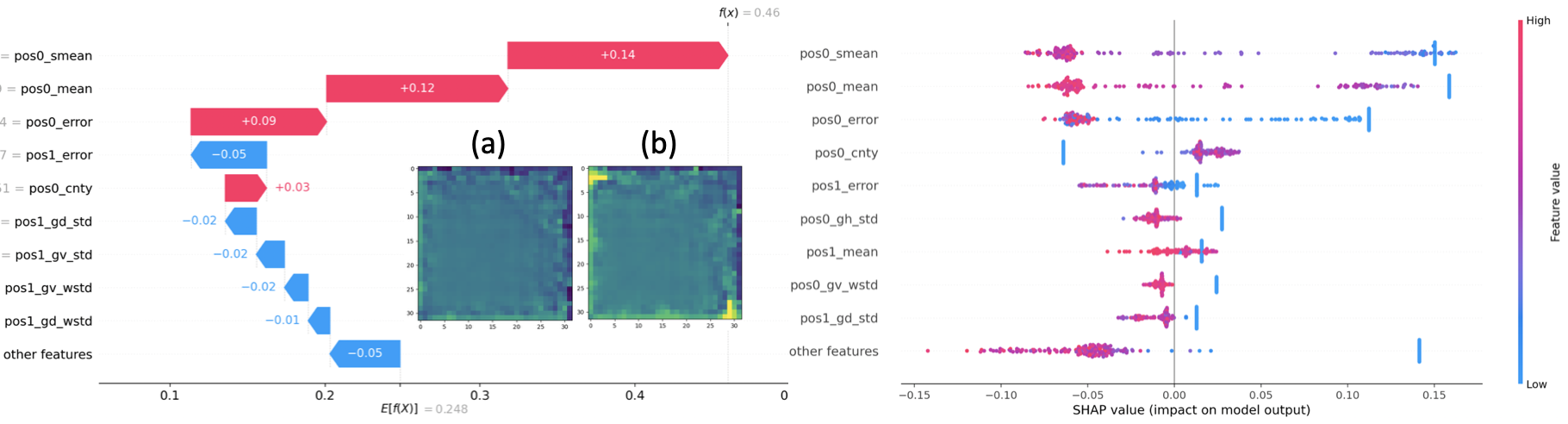}
    \caption{Shapley values of a randomly selected assessor (a random forest). Left: given a model $\mathrm{M_{5}}$, and its variance matrices generated at (a) the final prediction ($pos_0$) and (b) the final convolutional layer ($pos_1$), the assessor predicts the model to be ``background invariant". And the top reasons for this prediction is: the mean value of the variance matrix ($pos_0$) and the error rate (defined in \cite{ml4ml_invariance}) of the variance matrix ($pos_0$). Note that there are yellowish areas on the corners of the variance matrices at $pos_1$, therefore the continuity score and the gradient scores at these two positions have a negative impact on the decision. Right: In general, the assessor's decisions for being ``background-invariant" depend on the mean value of the variance matrix ($pos_0$) and the error rate of the variance matrix ($pos_0$), which is aligned with the analysis of $\mathrm{M_5}$ on the left.}
    \label{fig:shap}
\end{figure*}

In Table \ref{table: stat_acc} we show that the average accuracy of the models in the collection is around 91.9\%. We did not intentionally make any model not sufficiently trained. However, for some models that have not been pre-trained on ImageNet, the accuracy can drop to lower than 50\%.

In Figure \ref{fig:arch_stat}, we show that in our model repository, we have a wide range of different architectures, namely vision transformer patch16-224, resnet 18, and vgg variants. We also show the statistics for their background variance qualities. However, the size of our model collection is limited and we leave the studies of the architectures' impact on invariance qualities to future work.

\paragraph{C3. Analysis on Automated Assessor}

Shapley values (a game theoretic approach) \cite{shapley} are commonly used to explain outputs or predictions of machine learning models. It isolates each input feature and averages its expected marginal contribution. In this section, we report the usage of Shapley values to explain the predictions of the random forest (assessor) we train for prediction background invariance qualities of ML models.

As shown in Figure \ref{fig:shap}, for an arbitrarily selected model $\mathrm{M_{5}}$, the assessor's prediction is ``background invariant". And the top factor contributing to the predictions is the mean value of the variance matrix generated at $pos_0$ (final prediction). We also show that, in general, the assessor will consider the mean value of the variance matrix and the error rate (defined as the percentage of the pixels $\geq thres$ \cite{ml4ml_invariance}). This way, we can have a testing report for the assessor for both the overview analysis and case studies for any interested model(s).

\section{Variance Matrices from RBF-based Resampling.} 
Whilst scatter point clouds are able to display our testing results, they suffer from the problem of overlapping glyphs when the number of points per unit area becomes excessively large \cite{splatterplots}. Therefore, analysis and judgement based purely on scatter plots might not be consistent.

To address the problem of overlapping glyphs of scatter point clouds, we use the common approach of radial basis functions (RBFs) to transform a set of point clouds into a variance matrix. For each element $e$ in a variance matrix, an RBF defines a circular area in 2D, facilitating the identification of all data points in the circle. Let these data points be $p_1, p_2, \ldots, p_c$ and their corresponding values are $v_1, v_2, \ldots, v_c$. As discussed earlier, the coordinates of each data point are determined by the semantic distances from the target image to two testing images. A Gaussian kernel $\phi$ is then applied to these data points, and produces an interpolated value for element $e$ as 
\[
value(e) = \frac{%
\sum_{i=1}^{c} \bigl( \phi(\left \| e - p_i  \right \|) \cdot v_k \bigr)}{
\sum_{i=1}^{c} \phi(\left \| e - p_i \right \|)}
\]
However, when the RBF has a large radius, the computation can be costly. When the radius is small, there can be cases of no point in a circle. In order to apply the same radius consistently, we define a new data point at each element $e$ and 
use $K$ nearest neighbors algorithm to obtain its value $u(e)$. The above interpolation function thus becomes:
\[
value(e) = \frac{\phi(0) \cdot u(e) + 
\sum_{i=1}^{c} \bigl( \phi(\left \| e - p_i  \right \|) \cdot v_k \bigr)}{
\phi(0) + \sum_{i=1}^{c} \phi(\left \| e - p_i \right \|)}
\]

In Figure \ref{fig:SP2RBF}, we show the application of three different RBFs.
The mixed green and yellow patterns in row OL[1] gradually become more coherent towards OL[5].
We can clearly see a green square at the centre and yellow areas towards the top and right edges.

\section{Plausibility for Invariance Testing and Future Works} 
Consider two sets of testing images: X and Y with plausible and implausible background respectively. Model A performs 100\% correct with X, and 100\% incorrect with Y; Model B performs 100\% incorrect with X, and 100\% correct with Y. We should always prefer model A than B. For example, a case where a model fails to recognize a car just because it appears in front of a different building is a more serious issue than a case where it fails when the car appears in a bathroom. Therefore in this paper, we focus on invariance testing which takes plausibility into consideration. 
Figure \ref{fig:Similarity} shows three sets of example background scenes discovered for a targeting image (i.e., the fish image on the top-left corner of each set). 
While it is not necessary for every testing image in invariance testing to be realistic, the plausibility of a testing image reflects its probability of being captured in the real world.
It is unavoidable that invariance testing involves testing images of different plausibility, and therefore it is important to convey and evaluate the testing results with the information of the plausibility \cite{noise_bgtest, SCEGRAM}. An ideal set of background scenes should have a balanced distribution of scenes of different plausibility. Qualitatively, we can observe that in Figure \ref{fig:Similarity}, the random set has too many highly implausible images and the closest set has images biased towards keywords $K_x = \{$painting, water, tree$\}$, many are not quite plausible, while the expanded set has a better balance between more plausible to less plausible background scenes. When generating / synthesizing testing images, we adopted the simple background replacement in this work due to the reason stated in Section \ref{sec:simple_bg_replace}. In the future, this can be replaced by generative models when they become more reliable or when more advanced and accountable filtering techniques are available. Finally, in this work, for each target image, we select $N$ items from $K(\gg N)$ keywords based on semantic connections. In our approach, semantic connections depend on the frequency of association. In future work, more advanced techniques can be explored to improve the association analysis, e.g., a human-in-the-loop association mining technique.

\end{document}